\documentclass{article}

\usepackage{microtype}
\usepackage{appendix}
\usepackage{graphicx}
\usepackage{subfig}
\usepackage{booktabs} % for professional tables
\usepackage{MnSymbol,wasysym}
\usepackage{pifont}% http://ctan.org/pkg/pifont
\newcommand{\cmark}{\ding{51}}%
\newcommand{\xmark}{\ding{55}}%
\usepackage[subfigure]{tocloft}
\usepackage{etoolbox}
\usepackage{siunitx}
\usepackage{enumitem}
\setitemize{itemsep=0pt}
\usepackage{caption}
\usepackage[small,compact]{titlesec}
\newcommand{\gmark}{{\color{green}\cmark}}
\newcommand{\rmark}{{\color{red}\xmark}}

\usepackage{tikz}
\newcommand*\circled[1]{\tikz[baseline=(char.base)]{
            \node[shape=circle,draw,inner sep=2pt] (char) {#1};}}

\usepackage{hyperref}

\usepackage[accepted]{mlsys2022}

\mlsystitlerunning{MLPerf Mobile Inference Benchmark: An Industry-Standard Open-Source Machine Learning Benchmark for On-Device AI}

\begin{document}

\twocolumn[
\mlsystitle{MLPerf Mobile Inference Benchmark
}

\raggedbottom

% unless you've provided the [accepted] option to the mlsys2022


\mlsyssetsymbol{equal}{*}

\begin{mlsysauthorlist}
\mlsysauthor{Vijay	Janapa Reddi}{hu}
\mlsysauthor{David	Kanter}{mlc}
\mlsysauthor{Peter	Mattson}{g}
\mlsysauthor{Jared	Duke}{g}
\mlsysauthor{Thai	Nguyen}{g}
\mlsysauthor{Ramesh	Chukka}{i}
\mlsysauthor{Ken	Shiring}{mtk}
\mlsysauthor{Koan-Sin	Tan}{mtk}
\mlsysauthor{Mark	Charlebois}{qc}
\mlsysauthor{William	Chou}{qc}
\mlsysauthor{Mostafa	El-Khamy}{sam}\\
\mlsysauthor{Jungwook	Hong}{sam}
\mlsysauthor{Tom	St. John}{cru}
\mlsysauthor{Cindy	Trinh}{cindy}
\mlsysauthor{Michael	Buch}{hu}
\mlsysauthor{Mark	Mazumder}{hu}\\
\mlsysauthor{Relja	Markovic}{mlc}
\mlsysauthor{Thomas	Atta}{i}
\mlsysauthor{Fatih	Cakir}{sam}
\mlsysauthor{Masoud	Charkhabi}{g}\\
\mlsysauthor{Xiaodong	Chen}{sam}
\mlsysauthor{Cheng-Ming	Chiang}{mtk}
\mlsysauthor{Dave	Dexter}{arm}
\mlsysauthor{Terry	Heo}{g}\\
\mlsysauthor{Guenther	Schmuelling}{msft}
\mlsysauthor{Maryam	Shabani}{i}
\mlsysauthor{Dylan	Zika}{g}
\end{mlsysauthorlist}

\mlsysaffiliation{hu}{Harvard University}
\mlsysaffiliation{mlc}{MLCommons}
\mlsysaffiliation{g}{Google}
\mlsysaffiliation{i}{Intel}
\mlsysaffiliation{qc}{Qualcomm}
\mlsysaffiliation{arm}{ARM}
\mlsysaffiliation{cindy}{ENS Paris-Saclay}
\mlsysaffiliation{sam}{Samsung}
\mlsysaffiliation{msft}{Microsoft}
\mlsysaffiliation{mtk}{MediaTek}
\mlsysaffiliation{cru}{Cruise}

\mlsyscorrespondingauthor{Vijay Janapa Reddi}{vj@eecs.harvard.edu}
\mlsyscorrespondingauthor{David Kanter}{david@mlcommons.org}

\mlsyskeywords{Machine Learning, MLSys}

\vskip 0.2in

\begin{abstract}
This paper presents the first industry-standard open-source machine learning (ML) benchmark to allow performance and accuracy evaluation of mobile devices with different AI chips and software stacks. The benchmark draws from the expertise of leading mobile-SoC vendors, ML-framework providers, and model producers. It comprises a suite of models that operate with standard data sets, quality metrics and run rules. We describe the design and implementation of this domain-specific ML benchmark. The current benchmark version comes as a mobile app for different computer vision and natural language processing tasks. The benchmark also supports non-smartphone devices, such as laptops and mobile PCs. Benchmark results from the first two rounds reveal the overwhelming complexity of the underlying mobile ML system stack, emphasizing the need for transparency in mobile ML performance analysis. The results also show that the strides being made all through the ML stack improve performance. Within six months, offline throughput improved by 3$\times$, while latency reduced by as much as 12$\times$. ML is an evolving field with changing use cases, models, data sets and quality targets. MLPerf Mobile will evolve and serve as an open-source community framework to guide research and innovation for mobile AI.

\end{abstract}
]



\printAffiliationsAndNotice{} % otherwise use the standard text.

\section{Introduction}
\label{Intro}

Mobile artificial intelligence (AI)  applications are increasingly important as AI technology becomes a critical differentiator in smartphones, laptops, and other mobile devices.
%Many mobile applications benefit from AI: image processing, voice processing, and text interpretation. AI provides state-of-the-art solutions to  these tasks  with  a  quality that  users  will notice on  their  devices.  More and more  consumers  are  employing such applications, and they expect a high-quality experience---especially for applications with video or audio interactivity.
Consequently, laptops and smartphones have incorporated application-specific integrated circuits (ASICs) on the hardware front to support AI in an energy-efficient manner. 
%For machine learning, this situation leads to custom hardware that ranges from specialized instruction-set-architecture (ISA) extensions on general-purpose CPUs to ﬁxed-function accelerators dedicated to efﬁcient machine learning  (Figure~\ref{fig:combo}). Also, because mobile devices are complex, they incorporate a variety of features to be competitive, especially to conserve battery life.
% Artificial intelligence (AI) is emerging as a critical differentiator among mobile devices such as smartphones and laptops. Many consumer applications benefit from AI: image processing, voice processing, and text interpretation. AI provides state-of-the-art solutions to these tasks with a quality that users will notice on their smartphones and laptops. More and more consumers are employing such applications, and they expect a high-quality experience---especially for applications with video or audio interactivity. 
The software front includes many code paths and AI infrastructures to support machine-learning hardware efficiently. 

%Most System-On-Chip (SoC) vendors lean toward custom model compilation and deployment that integrates tightly with the hardware. Examples include Google's Android Neural Network API (NNAPI)~\cite{nnapi}, Intel's OpenVINO~\cite{OpenVINO}, MediaTek's NeuroPilot \cite{neuropilot}, Qualcomm's Snapdragon Neural Processing Engine (SNPE)~\cite{snpe} and Samsung's Exynos Neural Network (ENN) SDK~\cite{samsungsdk}. These frameworks handle different numerical formats (e.g., FP32, FP16, INT16, and INT8). They provide run-time support for various machine-learning networks that best fit the application when it is running on their hardware platform.

%Support for the technology is becoming common in nearly all mobile segments, from cost-optimized devices to premium phones. There are many AI approaches in practice, ranging from purely software-based techniques to hardware-supported machine learning that relies on tightly coupled libraries. 

\begin{figure}[t!]
        \includegraphics[width=\columnwidth]{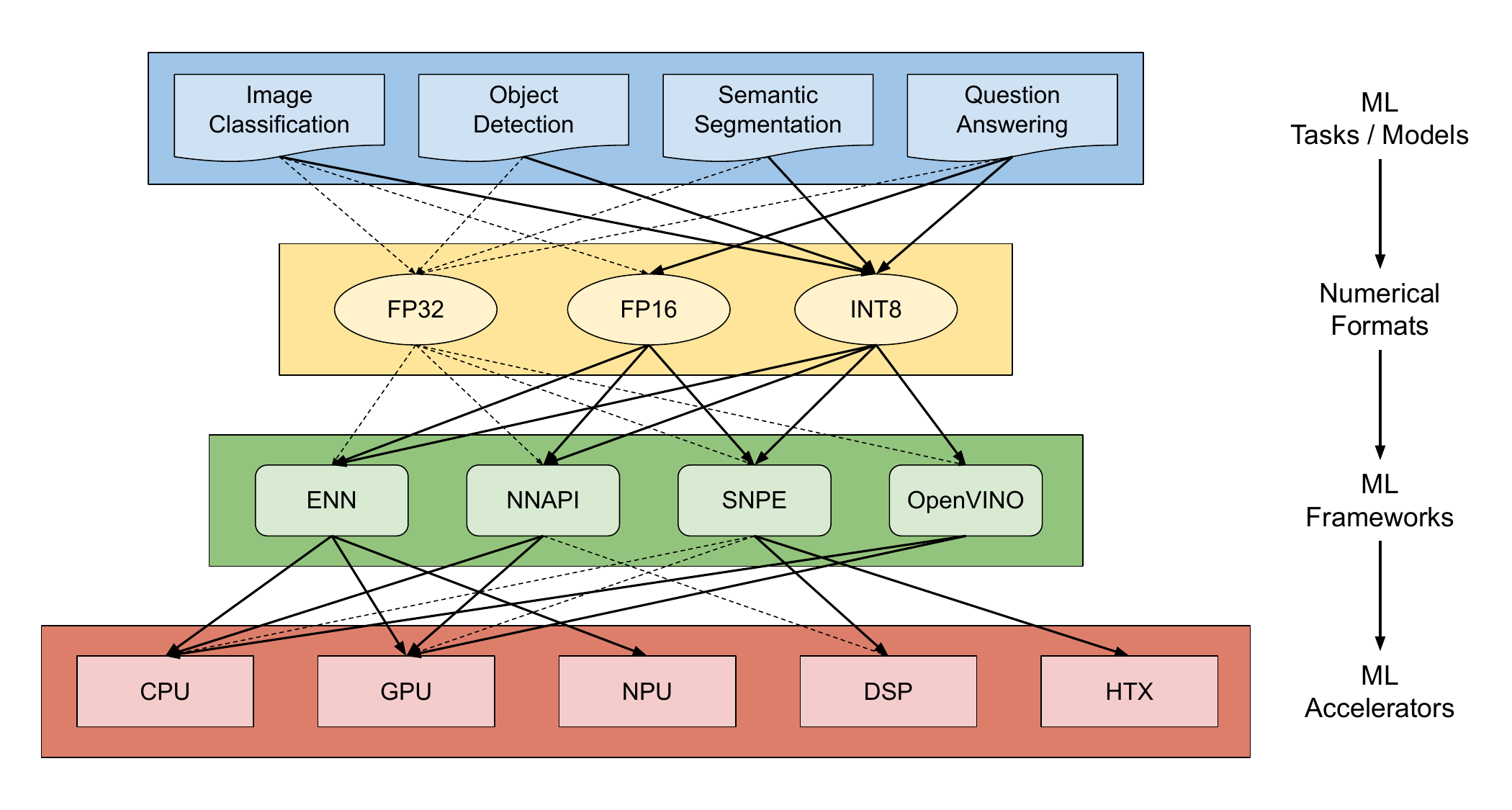}
        \vspace{-2em}
        \centering
        \caption{There are many ways to exercise a mobile SoC's rich suite of accelerators, which is why transparency is key.}
        \label{fig:combo}
        \vspace{-1em}
\end{figure}

While support for mobile AI applications is becoming a differentiating capability, seeing through the mist of competing solutions is difficult for fairly evaluating improvements in performance and efficiency. Figure~\ref{fig:combo} shows the number of different code pathways for generating results on mobile SoCs. The dashed lines represent mere possibilities, whereas the solid lines indicate actual submissions from various organizations. Different code paths yield different performance results. Therefore, benchmark-performance transparency is essential, as it reveals which code paths were taken, making the performance results reproducible and informative for consumers. OEMs, SoC vendors, researchers, and consumers can all beneﬁt when mobile devices employ AI in ways they can compare transparently and fairly. 

%For machine learning, this situation leads to custom hardware that ranges from specialized instruction-set-architecture (ISA) extensions on general-purpose CPUs to fixed-function accelerators dedicated to efficient machine learning. Also, because mobile devices are complex, they incorporate a variety of features to be competitive, especially those that conserve battery life. 

% are critical differentiators, many features employ application-specific hardware support. In machine learning, the result is custom hardware that ranges from specialized instruction-set-architecture (ISA) extensions in general-purpose CPUs to fixed-function accelerators dedicated to efficient processing. 

% As hardware and software support for mobile AI applications is becoming a differentiating capability, seeing through the mist of competing solutions is difficult for evaluating improvements in performance and efficiency. There is an increasing demand to make AI-performance evaluation transparent. OEMs, SoC vendors, researchers, and consumers benefit when mobile devices employ AI in ways they can see and compare. 

However, a key challenge for developing a robust mobile AI benchmark is, first and foremost, understanding the complex landscape of the mobile computing ecosystem. The end-user performance of a mobile AI device is more than its AI hardware capability in isolation. Instead, a more accurate measurement results from the AI hardware coupled with its ML-software framework, whose net performance is shrouded beneath layers of developer options, deployment scenarios, and the OEMs' lifecycles. This complexity is not discussed nor relevant for major server-side inference benchmarks~\cite{gao2019aibench,reddi2020mlperf}. As such, there is a need for a domain-specific benchmark that can critically compare and evaluate systems with mobile-specific models, numerics, frameworks, metrics, and methodology.

To address the challenge, we take an open-source, community-driven approach. A consortium of mobile vendors and academic organizations with shared interests, yielding collective expertise in mobile neural-network models, data sets, and submission rules, have developed the MLPerf Mobile benchmark to ensure the results are relevant to the industry and academia~\cite{MLCmobile}, and beneficial to consumers while being transparent and reproducible. The following five principles inform the benchmark's design:
\begin{enumerate}
  \setlength\itemsep{0em}
    \item The benchmark must capture the \textbf{real-world mobile system complexity} involved in delivering AI performance to users who procure a commercial device. 
    % We want to prevent the benchmark from implementing special code beyond what these users generally employ.
    \item The benchmark must \textbf{identify mobile-specific models and represent diverse tasks} that are challenging and resist model- and domain-specific optimizations.
    \item Each task should have an \textbf{appropriate accuracy and minimum quality threshold} that matches and reflects the metrics that matter for mobile AI device end-users.
    \item The \textbf{testing conditions must closely match the environments in which mobile devices typically serve}, such as ambient temperature and battery power.
    \item Performance \textbf{results must be publicly reproducible outside the submitting organization} as commercial mobile devices are globally accessible and to foster generational ML advancements on prior achievements. 
\end{enumerate}

Our approach to addressing principles 1--5 was to develop an industry-neutral open-source MLPerf Mobile app that all benchmark tests must use. The initial version has a set of four mobile-specific neural network models for three computer vision (CV) tasks and one natural language processing (NLP) task, each with its own accuracy and minimum quality targets. The app passes these models to the back-end layer, which is an abstraction that allows different hardware (and software) vendors to optimize their implementations for neural networks. The app also has a presentation layer for wrapping the more technical benchmark layers and the Load Generator (``LoadGen''). The LoadGen allows representative testing of different inference platforms and use cases by generating inference requests in a pattern and measuring specific parameters (e.g., latency, throughput, or latency-bounded throughput). The benchmark also offers a headless version of the mobile application that enables laptops running non-mobile OSs to use the same benchmarks.

Two rounds of MLPerf Mobile submissions have been completed~\cite{v10}. Comparing the results between these two generations reveals these key takeaways:
        \vspace{-2em}

\begin{itemize}
  \setlength\itemsep{0em}
    \item \textit{Benchmarking drives generational improvements}. Between two versions of the benchmark (six months), latency improved by 2$\times$ on average and by 12$\times$ in one case. Developing principled methods to measure mobile ML performance is important to drive innovation. 

    \item \textit{There is no one size fits all}. The results show a range of hardware and software approaches to implement neural network models efficiently on mobile devices. The approach needs to be driven by the app use case. 

    \item \textit{Accelerator level parallelism (ALP) is important}. Vendors exercise multiple hardware accelerators concurrently to maximize offline throughput performance. Therefore, there is a need for managing hardware ALP.

    \item \textit{State-of-the-art should compare against vendor-backends.} Mobile AI accelerators often rely on vendor-specific SDKs and custom backends to unleash their full potential as more general-purpose frameworks like NNAPI can lead to 10\% slower performance, or worse, be 7$\times$ slower due to buggy support~\cite{buch2021ai}.

    \item \textit{Numerics (still) matter.} Not all mobile ML tasks benefit from INT8 quantization. Tasks like NLP require FP16 arithmetic to be useful in real deployments, implying not everything needs a dedicated AI accelerator.

        % \item \textit{Performance transparency is not optional, it is needed.} The mobile ecosystem is highly fragmented. It requires a deep understanding of the ecosystem and a strong commitment to fairness and reproducibility so that we can have better result transparency.
    
\end{itemize}

% MLSys is emerging as a flagship venue for ML systems research. However, despite the widespread use of mobile ML systems, there is a lack of awareness and emphasis about these systems in MLSys. We hope that the industry approach we present, sourced from ten major and leading hardware and software organizations, provides architects, ML system designers and developers insights into the challenges, issues, and opportunities in assessing mobile ML. 

Ideally, researchers would track MLPerf~Mobile's benchmark tasks, accuracy metrics, quality thresholds, rules, etc., to present industry-relevant evaluations that practitioners can adopt to bridge the gap between research and practice. As the mobile AI landscape is vastly different from desktop and cloud AI deployments, our open-source mobile app can be a common baseline for integrating various ML frameworks and models, facilitating ``out of the box'' research needed for reproducibility on real devices and  simulators. Supported by MLCommons~\cite{mlcommons}, MLPerf Mobile will continue to evolve and stay up-to-date.

\section{Mobile AI Ecosystem Challenges}
\label{Benchmarking Challenges}

%We discuss challenges that are unique to mobile computing and do not arise in traditional server, edge and desktop computing systems. 

Mobile AI performance is shrouded behind multiple layers of complexity. We describe these important factors that significantly impact mobile AI performance in the real world: hardware heterogeneity, software fragmentation, developer options, deployment scenarios, and OEM life cycles. Each by itself leads to performance variability, but the combination makes AI benchmarking extremely challenging. 

%Figure~\ref{fig:nested hierarchy} shows the various constituents and explains the implementation options and challenges facing each one.
% \begin{figure}[t!]
%         \includegraphics[width=0.85\columnwidth]{fig/nested_hierarchy.pdf}
%         \centering
%         \caption{Mobile AI performance constituents.}
%         \label{fig:nested hierarchy}
% \end{figure}

\subsection{Hardware Heterogeneity}
\label{Hardware Heterogeneity}

A given device may have a spectrum of AI-performance capabilities, depending on which processing engines it uses. Smartphones contain complex heterogeneous chipsets that provide many different compute units and accelerators. A typical mobile system-on-a-chip (SoC) complex includes a CPU cluster, GPU, DSP, neural processing unit (NPU), Hexagon Tensor Accelerator (HTA), Hexagon Vector Extensions (HVX), and so on. Any or all of these components can aid in machine-learning (ML) inference. Moreover, many smartphones today are Arm-based, but the CPU cores generally implement a heterogeneous ``big.LITTLE'' architecture \cite{biglittle}. Some SoCs even have big-CPU clusters where some CPUs clock faster than others. Also, devices fall into different tiers with different hardware capabilities at different prices, varying in their memory capacity and storage features. Any processing engine can run ML workloads, but this flexibility also makes benchmarking AI performance difficult. Hence, there is a need for a transparent way to benchmark a smartphone's AI-hardware performance. 

%% Lots of places where we can run the ML code, CPU, GPU, DSP, NPU, AIP, etc. in some cases multiple flavors of each (e.g., big and little cores for CPU / AIP)

\begin{figure}[t!]
\vspace{-12pt}
        \subfloat[]{\includegraphics[width=0.3\columnwidth]{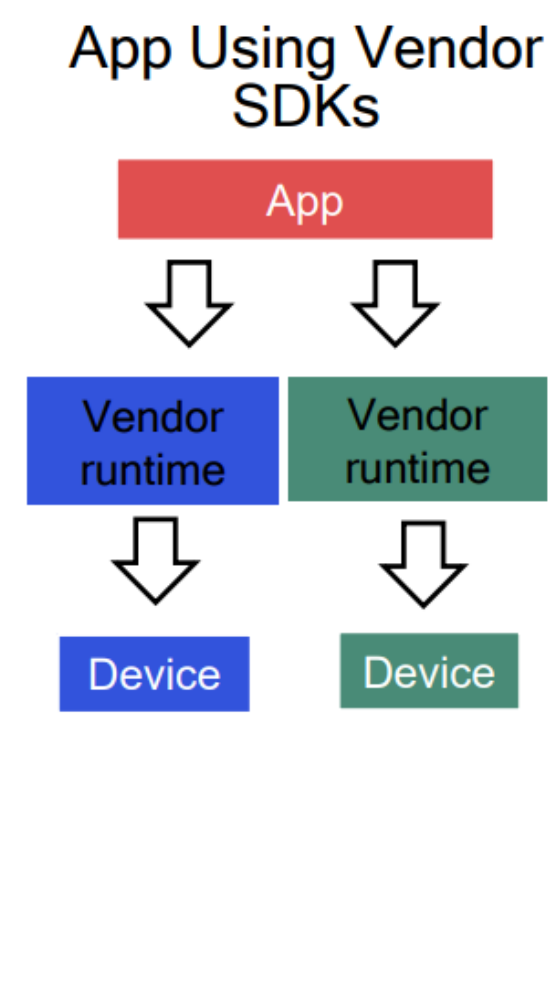}\label{fig:app_dev_option_a}}
        \subfloat[]{\includegraphics[width=0.3\columnwidth]{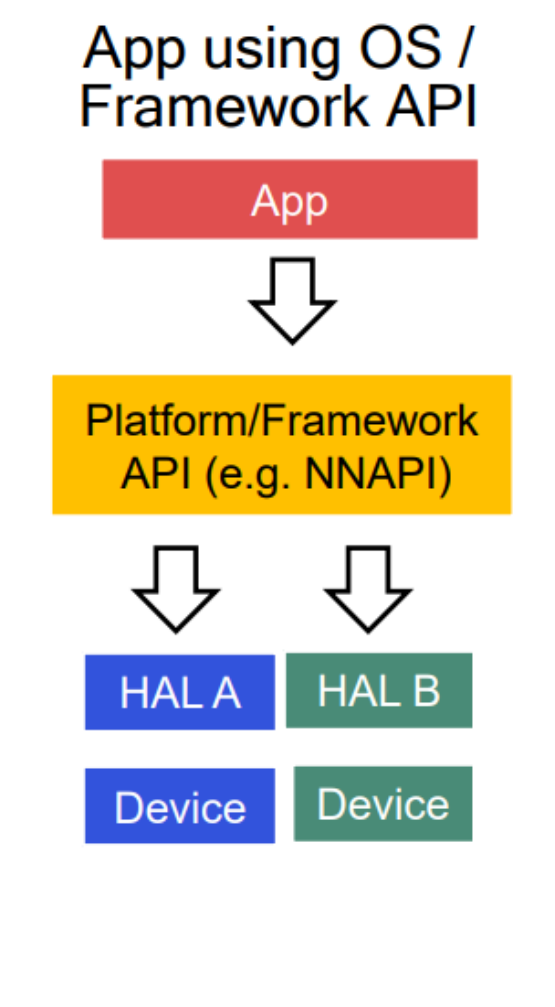}\label{fig:app_dev_option_b}}
        \subfloat[]{\includegraphics[width=0.3\columnwidth]{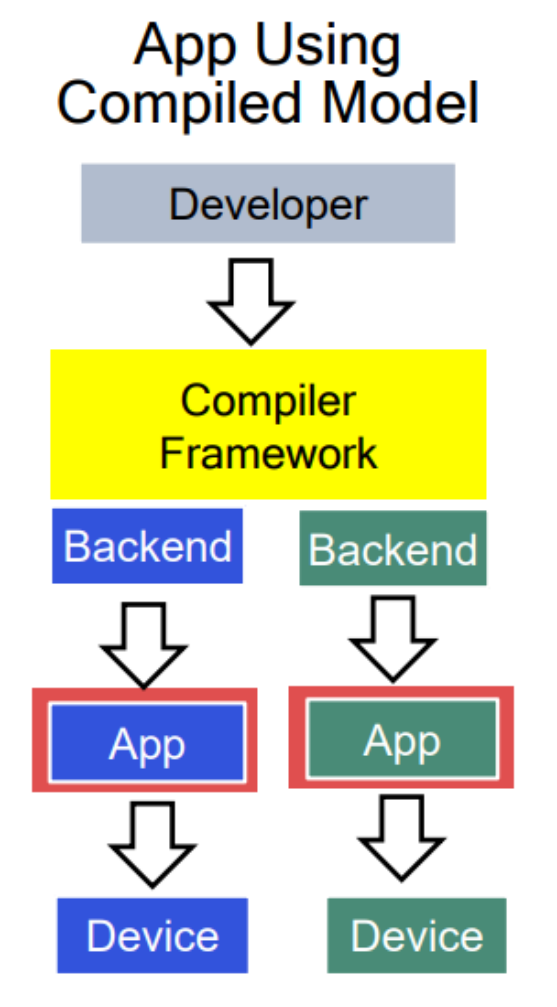}\label{fig:app_dev_option_c}}
        \centering
        \caption{Application-development options.}
        \label{fig:app_dev_option}
\vspace{-15pt}
\end{figure}

\subsection{Software Fragmentation}
\label{Software Fragmentation}
%% what makes software runtimes complicated on mobile? Show the 3 code paths possible here. 1st party apps using vendor SDKs, 3rd party optimized apps which optimize for a platform using vendor SDKs, 3rd party portable apps using NNAPI.code paths that are possible here.

%%% I vaguely remember a presentation by Mark that covers the SW fragmentation
The mobile software ecosystem is heavily differentiated from the OS to the run-time ML framework. The diversity of different software code paths can drastically affect hardware performance. Hence, a transparent mechanism for operating, introspecting, and evaluating a mobile device is essential. 

Mobile devices employ various OSs: Android, iOS, Windows, Ubuntu, Yocto, etc. Each OS has an ecosystem of ML application programming interfaces (APIs) and application-deployment options that necessitate particular solutions. Numerous APIs have served in the development of ML applications. A single SoC or OEM device will often have to support a vendor SDK and/or a plurality of frameworks. 

\begin{figure}[t!]
\vspace{-14pt}
        \subfloat[]{\includegraphics[width=0.3\columnwidth]{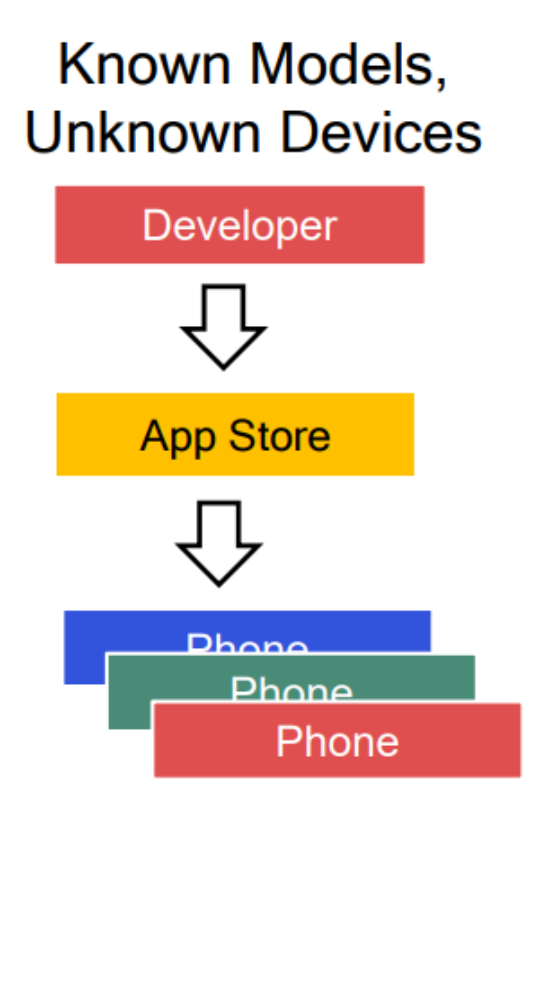}}
        \subfloat[]{\includegraphics[width=0.3\columnwidth]{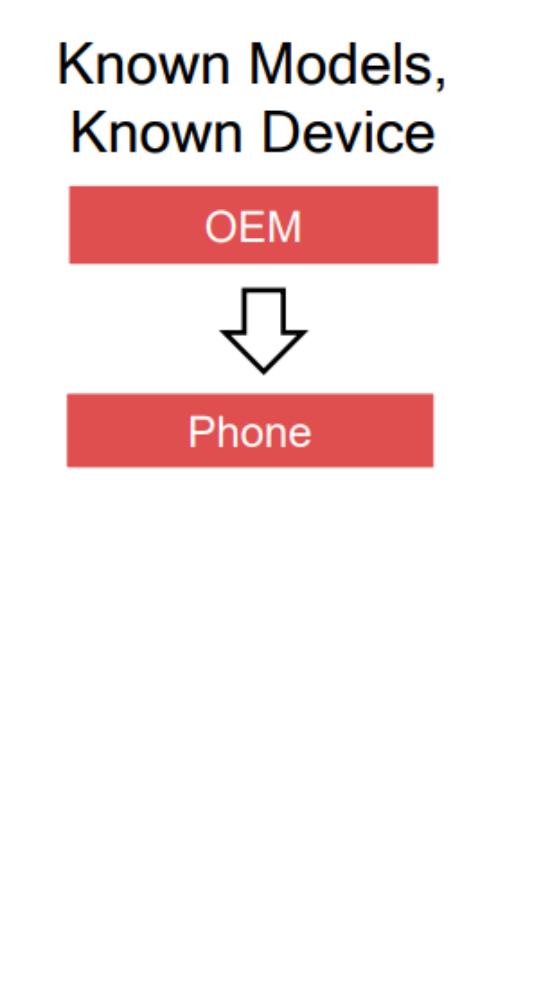}}
        \subfloat[]{\includegraphics[width=0.3\columnwidth]{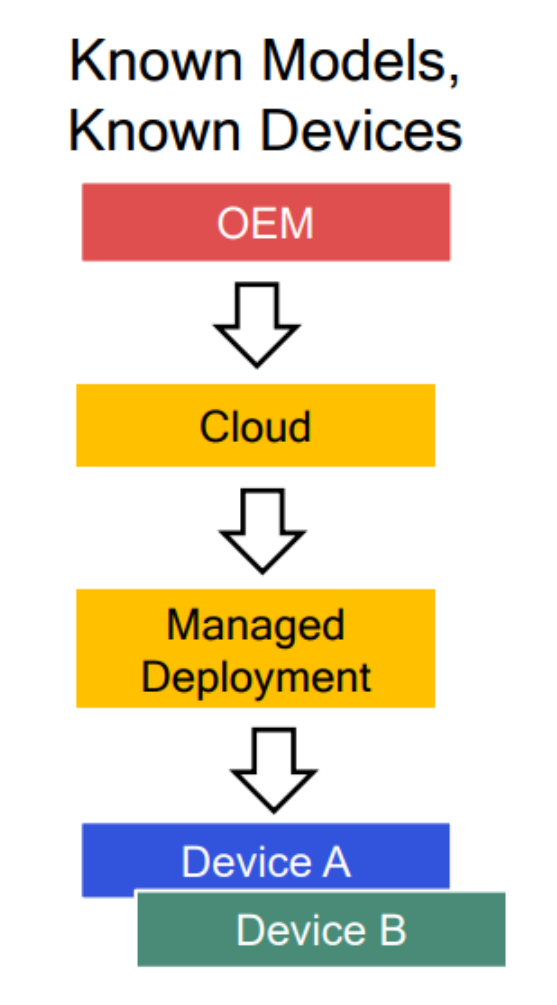}}
        \centering
        \caption{ML-application deployment scenarios.}
        \label{fig:deployment scenarios}
\vspace{-15pt}
\end{figure}

SoC vendors offer a proprietary software development kit (SDK) that generates optimized binaries so ML models can run on SoC-specific hardware. These vendors also make engineering investments to support more generic frameworks, such as TensorFlow Lite (TFLite) \cite{tensorflowlite} and NNAPI \cite{nnapi}, that provide a compatibility layer to support various accelerators and device types. But as resources are limited, SoC vendors prioritize their SDKs, resulting in less-optimal generic-framework support.

The diversity of vendor SDKs and framework-support levels are all reasons why the mobile-ML software ecosystem is fragmented. This situation complicates hardware-performance assessment because the choice of software framework has a substantial effect. A high-performance SoC, for instance, may deliver low performance, owing to an ill-matched framework. For example, even if a SoC integrates a high-performance ML accelerator, if a generic Android framework like NNAPI does not support it with high-performance driver back ends, the accelerator will function poorly when handling a network~\cite{buch2021ai}. 

\subsection{Developer Options}

The ecosystem allows application developers to choose among several different approaches to enable machine learning on mobile devices, making it necessary to have an open-source methodology for understanding performance. Application developers can work through a marketplace such as Google Play \cite{googleplay} to create mobile-app variants for every SoC vendor if they follow a vendor-SDK approach (Figure~\ref{fig:app_dev_option_a}). However, doing so presents a scalability challenge because of the increased time to market and additional development costs. An alternative is to create an application using a native OS/framework API such as NNAPI, which provides a more scalable approach (Figure~\ref{fig:app_dev_option_b}). Nevertheless, this alternative has a crucial shortcoming: it is only viable if SoC vendors provide good back-end drivers to the framework, necessitating cooperation between them and the framework designers. A final alternative is to bind the neural-network model to the hardware. Doing so allows compilation of the model to a particular device, avoiding reliance on any particular run time (Figure~\ref{fig:app_dev_option_c}), but this lacks device portability which is needed for mobile computing.

%% talk about the mobile device heterogeneity. Say how many devices are involved etc. and so you can’t have a magic silver bullet app. Got to make some tough touch decisions on backward compatibility etc. 

%% system cache, DRAM differences, system-level differences and optimization

%%% Machine Learning at Facebook: Understanding Inference at the Edge could be used here [Cite]

%%% Mark’s presentation could be used here as well

\subsection{Deployment Scenarios}

Mobile ML applications have many potential uses. Details of the use case determine the extent to which an ML model is optimized for the hardware and how it runs, because of strong or weak ties to the device. Developers primarily build applications without specific ties to vendor implementations. They may design custom neural-network models that can run on any device. Thus, mobile devices often run apps that employ unknown models for a variety of hardware (Figure~\ref{fig:deployment scenarios}(a)). OEMs, on the other hand, build their ML applications for their own devices. Therefore, both the models and the device targets are known at deployment time (Figure~\ref{fig:deployment scenarios}(b)). A service provider (e.g., Verizon or AT\&T) that uses a variety of hardware solutions may, however, support its service with known models, in which case both the models and the hardware are known (Figure~\ref{fig:deployment scenarios}(c)). Development of the applications deployed in these scenarios may also take place in various ways. OEMs that manufacture devices can use vendor SDKs to support their applications with minimal extra effort. Given these options, it is necessary to know which of these underlying approaches is being used to produce the measured AI performance results.

%% from a benchmarking standpoint, talk about how the mobile device lifecycle works because this has implications on performance benchmark. For instance, we have NNAPI release cycles that we need to align with. OEM control over SW stack, etc.

\subsection{OEM Lifecycle}

%Because of the way commercial mobile devices (particularly smartphones) operate, getting reproducible numbers can be difficult. 

%When benchmarking a device for performance, a newly installed original equipment manufacturer (OEM) software update may affect the results, and installing the same version of the software used to generate a particular benchmark result may be impossible. After a device applies a system-software update, the only way to revert to the previous configuration is to factory reset the device. But doing so also undoes any associated security fixes.

%Mobile-SoC testing often occurs on development platforms. Gaining access to them, however, is difficult. Therefore, the results of benchmark testing that employs a development platform may not be independently verifiable. For this reason, benchmarking generally takes place on commercial devices.

A variety of other factors, ranging from how OEMs package software for delivery to how software updates are issued, also affect hardware-performance measurements. OEMs employ vendor SoCs and associated software releases to produce commercial mobile devices. 
%In the case of smartphones, those devices may sell unlocked or locked to a wireless carrier, in which case the carrier ultimately controls the software. If the carrier sells the device, it will likely require testing and validation before allowing any device updates. All of these factors can lead to delays to the software-update channel. 
OEMs pick up the software updates (such as framework enhancements) from the SoC vendors and bundle them with other updates for periodic release. Usually, a delay occurs between the time when an SoC vendor releases a software update and when that performance update sees deployment. The delay is months long. Moreover, commercial devices receive OEM updates only for a fixed period, so they will not benefit from software-performance enhancements afterward. Thus, getting reproducible numbers is difficult without transparency.

\renewcommand{\arraystretch}{1}

\begin{table*}[t!]
\resizebox{\linewidth}{!}{
\begin{tabular}{|l|l|l|l|l|l|}
\hline
\multicolumn{1}{|c}{\textbf{Version}}  & \multicolumn{1}{|c}{\textbf{Area}}     & \multicolumn{1}{|c}{\textbf{Task}}                        & \multicolumn{1}{|c}{\textbf{Reference Model}}                   & \multicolumn{1}{|c}{\textbf{Data Set}}                                    & \multicolumn{1}{|c|}{\textbf{Quality Target}}                        \\ \hline
v0.7, v1.0 & Vision   & Image classification        & MobileNetEdgeTPU (4M params)   & ImageNet 2012 (224x224) & 98\% of FP32 (76.19\% Top-1) \\ \hline
v0.7, --- & Vision   & Object detection             & SSD-MobileNet v2 (17M params) & COCO 2017 (300x300)      & 93\% of FP32 (24.4\% mAP)              \\ \hline
---, v1.0 & Vision   & Object detection             & MobileDET-SSD (4M params) & COCO 2017 (320x320)      & 95\% of FP32 (28.5\% mAP)              \\ \hline
v0.7, v1.0 & Vision   & Semantic segmentation & DeepLab v3+ (2M params)        & ADE20K (512x512)        & 97\% of FP32 (54.8\% mIoU)            \\ \hline
v0.7, v1.0 & Language & Question answering          & MobileBERT (25M params)        & Mini Squad v1.1 dev                           & 93\% of FP32 (93.98\% F1)                \\ \hline
\end{tabular}
}
\caption{MLPerf Mobile benchmark suite. In the second version (v1.0) of the benchmark, we updated the object detection model to be more representative of industry needs and this is also reflected in the more stringent quality target requirements.}
\label{tab:task_models}
\vspace{-1em}
\end{table*}

\section{MLPerf Mobile Benchmarks}
\label{MLPerf Mobile Benchmarks}
% MLPerf Mobile Inference is community-driven. As such, all involved parties aided in developing the benchmark models and submission rules; the group includes both submitting organizations and organizations that care about mobile AI. Participants reached a consensus on what constitutes a fair and useful benchmark that accurately reflects mobile-device performance in realistic scenarios. 

% Deep learning inference spans a broad spectrum of devices, ranging from datacenter scale multi-node systems to tiny IoT endpoint devices that consume only milliwatts of power. Each of these deployment scenarios comes with its unique constraints and challenges (as discussed in Section~\ref{Benchmarking Challenges}). For example, cloud-scale systems are focused on high-performance and the total cost of operation (TCO). In contrast, mobile devices are resource-constrained, lack much memory and battery capacity, and face tight thermal envelopes. Thus, despite the adoption of methods to measure inference performance using MLPerf~\cite{reddi2020mlperf}, there is a need for more domain-specific benchmarks that can compare and evaluate systems with domain-specific models, frameworks, and hardware. 

To tackle the challenges, we developed MLPerf Mobile~\cite{MLCmobile}. A key aspect of our work is the methodology more so than the specifics of a benchmark version.

\subsection{Benchmark Design Philosophy}
\label{sec:phil}

The ML landscape is evolving and there are a plethora of models in the wild. Mobile ML systems include a wildly wide range of use cases, ranging from light-weight (e.g., classification on small 300x300 pixel images) to heavy-weight (e.g., 50~MP camera on Xiaomi for 5$\times$ zoom) tasks. Grappling with the rich diversity of tasks and models requires a long-term perspective focused on building a robust benchmark. In the initial version, we intentionally selected a few machine-learning tasks that represent light- and mid-weight mobile use cases that are stable, rather than picking models that are still evolving where there is no agreement on which mobile ML model versions are broadly applicable. We chose networks for a small set of tasks based on their maturity and applicability to different hardware (CPUs, GPUs, DSPs, NPUs, etc.). As we show later in the evaluation section, benchmarking these initial models still yields helpful insights about hardware performance across various deployment scenarios. We discuss and present our plans to extend the benchmark suite tasks over time in Appendix~\ref{Future work}.

\subsection{ML Tasks and Models}
\label{Tasks and Models}

% Machine-learning tasks and associated neural-network models come in a wide variety. Rather than support numerous models, however, our initial versions of the benchmark focus on establishing a high-quality benchmarking method. 

\textbf{Image classification.} We selected MobileNetEdgeTPU~\cite{Howard2019}, a well-optimized mobile model that usually provides good performance on different SoCs. MobileNetEdgeTPU is a descendent of the MobileNet-v2 family optimized for low-latency and mobile accelerators. The architecture is based on convolutional layers with inverted residuals and linear bottlenecks, similar to MobileNet v2. Still, it is optimized by introducing fused inverted bottleneck convolutions to improve hardware utilization and by removing hard-swish and squeeze-and-excite blocks. Evaluation of the MobileNetEdgeTPU reference model employs the ImageNet 2012 validation data set \cite{ILSVRC15} and requires 74.66\% (98\% of FP32 accuracy) Top-1 accuracy. Before inference, images are resized, cropped to 224x224, and then normalized.

%picks the best label to describe an input image and commonly serves in photo search, text extraction, etc. Many mobile applications (e.g., Google Lens~\cite{GoogleLe77:online}) employ classification to first identify relevant objects. Classifier-network evaluation is a good performance indicator when it serves as a feature-extractor backbone for other tasks. 

% This image classification task uses the ImageNet 2012 validation data set \cite{ILSVRC15}. Before inference, images are resized, cropped to 224x224, and normalized. We evaluated the model using the Top-1 accuracy metric.

\textbf{Object detection.} Our v0.7 reference model is the Single Shot Detector (SSD) \cite{liu_2016} with a MobileNet v2 backbone \cite{sandler2019mobilenetv2}---a choice that is well adapted to constrained computing environments. SSD-MobileNet v2 uses MobileNet v2 for feature extraction and uses a mobile-friendly SSD variant called SSDLite~\cite{sandler2019mobilenetv2} for detection. SSD prediction layers replace all the regular convolutions with separable convolutions (depthwise followed by 1x1 projection). SSD-MobileNet v2 reduces latency by decreasing the number of operations; it also reduces the memory that inference requires by never fully materializing the large intermediate tensors. Two SSD-MobileNet v2 versions acted as the reference models for the object-detection benchmark, one model replacing more of the regular SSD-layer convolutions with depth-separable convolutions than the other does. We used the COCO 2017 validation data set \cite{lin2015microsoft} and, for the quality metric, the mean average precision (mAP). The target accuracy is an mAP value of 22.7 (93\% of FP32 accuracy). The preprocessing stage resizes the image to 300x300---typical of resolutions in smartphones---and then does normalization.

In v1.0, we updated the reference model to MobileDets~\cite{xiong2021mobiledets} with the SSDLite that is more geared toward stressing mobile hardware accelerators such as mobile CPUs, GPUs, EdgeTPUs and DSP. A key feature of MobileDets is that in addition to using inverted bottlenecks as the only building block, it injects regular convolution operations into the neural network. Regular convolutions help improve the accuracy-latency trade-off on several hardware accelerators when placed at the appropriate positions in the network. We continue to use COCO 2017 validation set for testing with mAP as the quality metric. The image input size is different in MobileDets. It increases the input image resolution from 300x300 to 320x320. MobileDets has fewer parameters than MobileNets v2, but the high image resolution demands increased computation.

%draws bounding boxes around objects in an input image and labels those objects, often in the context of camera inputs. It has several use cases in mobile, especially in mobile-retail tasks for identifying items in a picture.  Implementations often use a pretrained image-classifier network as a backbone or feature extractor, then perform bounding-box selection and regression for precise localization \cite{ren2016faster, liu_2016}. 

\textbf{Semantic image segmentation.} Our reference model for this task in the first version is DeepLab v3+ \cite{chen2018encoderdecoder} with a MobileNet v2 backbone. DeepLab v3+ originates from the family of semantic image-segmentation models that use fully convolutional neural networks to directly predict pixel classification \cite{long2015fully, eigen2015predicting} as well as to achieve state-of-the-art performance by overcoming reduced-feature-resolution problems and incorporating multiscale context. It uses an encoder/decoder architecture with atrous spatial pyramid pooling and a modular feature extractor. We selected MobileNet v2 as the feature extractor because it enables state-of-the-art model accuracy in a constrained computational budget. We chose the ADE20K validation data set \cite{zhou2017scene} for its realistic scenarios, cropped and scaled images to 512x512, and (naturally) settled on the mean intersection over union (mIoU) for our metric. Additionally, we trained the model to predict just 32 classes (compared with 150 in the original ADE20K data set); the 1st to the 31st are the most frequent (pixel-wise) classes in ADE20K, and the 32nd represents all the other classes. The mIoU depends on the pixels whose ground-truth label belongs to one of the 31 most frequent classes, boosting its accuracy by discarding the network's bad performance on low-frequency classes.

\textbf{Question answering.} We selected MobileBERT \cite{sun2020mobilebert}, a BERT model that is well suited to resource-limited mobile devices. Further motivating this choice is the model's state-of-the-art performance and task-agnostic nature: even though we consider question answering, MobileBERT is adaptable to other NLP tasks with only minimal fine-tuning. We trained the model with a maximum sequence length of 384 and use the F1 score for our metric. This task employs the Stanford Question Answering Dataset (Squad) v1.1 Dev \cite{rajpurkar2016squad}. Given a question and a passage from a Wikipedia article, the model must extract a text segment from the passage to answer the question.

\textbf{Others.} Our current network choices reflect common usecases. Most mobile ML use cases involve computer vision. But building on our design philosophy (Section~\ref{sec:phil}), Appendix~\ref{Future work} discusses our plans to extend the benchmark. For example, a mobile version of RNN-T for speech is in the works and we will continue to add new models in the future.

%is an NLP task where the system must respond to human-posed questions in colloquial language. Today's device users get information through various mobile applications, including browsers, email, social media, and others. Question-answering across all this data accessed on a device can aid users in finding information quickly.

%partitions an input image into labeled objects at pixel granularity. It has many use cases in mobile applications, such as background replacement, photo processing, person segmentation, and AR/VR.
%to autonomous driving and robotics  \cite{8436956,9026297, neverova, siam2017deep}, remote sensing \cite{sherrah2016fully}, medical imaging \cite{taghanaki2020deep}, and complex image manipulation such as red-eye reduction.

% Example applications include search engines, chatbots, and other information-retrieval tools. Recent NLP models that rely on pretrained contextual representations have proven useful in diverse situations \cite{dai2015semisupervised, peters2018deep,radford2018improving}. BERT (Bidirectional Encoder Representations from Transformers) \cite{devlin2019bert} improves on those models by pretraining the contextual representations to be bidirectional and to learn relationships between sentences using unlabeled text. 

%%%%%%%%%%

%% screenshot of the application so people get an intuitive feel. This would likely be a two column figure that shows all the wireframes that audrey shares, finally.

\subsection{Reference Code}
\label{Reference Code}
%% explain what are the set of principles we wanted to follow to build the application. What are these by the way? I can’t recall if we had a principled approach :)
We provide reference-code implementations in the TensorFlow and TensorFlow Lite (TFLite) formats. We choose TF because it is vendor-neutral and most vendor backends can easily import TF code and models into their internal optimization frameworks. The reference code's goal is to identify the critical model-invocation stages. For instance, the reference benchmarks implement the preprocessing steps and the model's input-generation procedure. Benchmark submitters may readily adopt the code for their submission. Or they may choose to optimize these stages (e.g., rewrite them in C instead of Python) for performance---as long as they employ all the same stages and take the same steps to maintain equivalence. All reference models have 32-bit floating-point weights, and the benchmark additionally includes an 8-bit quantized version (with either post-training quantization or quantization-aware training, depending on the tasks). The reference implementations are available as open-source and free for users to download \cite{mlperf}.

The reference implementation is poorly optimized. Vendors that submit results to MLPerf must inherit the reference code, adapt it, and produce optimized glue code that performs well on their hardware. For example, to handle (quantized) inference, they need to invoke the correct software back end (e.g., SNPE or ENN) or an NNAPI driver to schedule code for their SoC’s custom hardware accelerators. 

\section{Load Generator}
\label{System Under Test}

This section describes how we generate load onto the system under test (SUT) and measure inference performance. 

%Mobile-SoC testing often occurs on development platforms. However, gaining access to them is difficult. Therefore, benchmarking generally takes place on commercial devices. 
% A system under test (SUT) interfaces with several device components. Orchestrating the  SUT execution involves multiple stages. The main ones are model selection, data-set input, preprocessing, back-end execution, and postprocessing. Figure~\ref{fig:Load generator} shows how they work together. The first step is reference-model selection: either TensorFlow or TFLite. 

\subsection{LoadGen}

To enable testing of various inference platforms and use cases, we devised the Load Generator (``LoadGen'')~\cite{github}, which creates inference requests in a pattern and measures some parameters (e.g., latency, throughput, or latency-bounded throughput). It logs information about the system during execution to enable post-run validation. Submitter modification of the LoadGen software is forbidden to ensure the testing behavior remains intact.

\begin{figure}[t!]
        \includegraphics[width=\columnwidth]{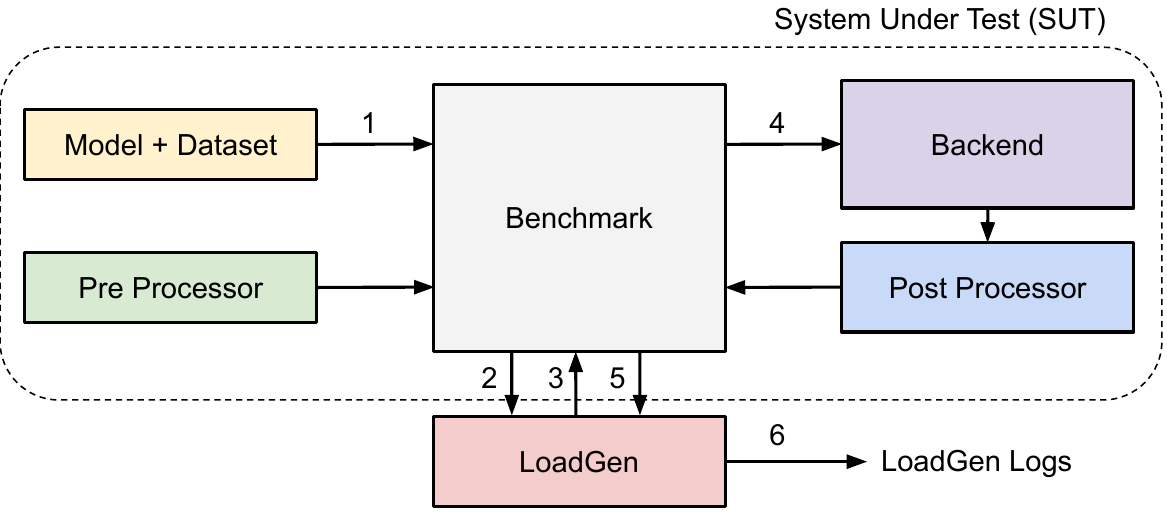}
        \centering
        \caption{Load Generator (``LoadGen'') testing the SUT.}
        \label{fig:Load generator}
\vspace{-15pt}
\end{figure}

As Figure~\ref{fig:Load generator} shows, the LoadGen uses the data sets as inputs to the SUT. It feeds the entire data set to the SUT to verify that the model delivers the required accuracy in accuracy mode. It provides only a subset of images to the SUT to measure steady-state performance in performance mode. A seed and random-number generator allows the LoadGen to select samples from the data set for inference, precluding unrealistic data-set-specific optimizations.

For pre-processing, the typical image-preprocessing tasks---such as resizing, cropping, and normalization---depend on the ML model. This stage implements data-set-specific preprocessing that varies by task, but all submitters must follow the same steps. Post-processing is a data-set-specific task that covers all the ops needed for accuracy calculations. For example, computing the Top-\textit{N} results for an image classifier requires a Top-K op / layer after the softmax layer. 

The backend for the reference benchmark implementation (Section~\ref{Reference Code}) is a TFLite smartphone back end that optionally includes NNAPI and GPU delegates. We also provide a ``dummy'' back end as an example reference for proprietary back ends; submitters replace it with whatever corresponds to their system. For instance, Qualcomm would replace the dummy with SNPE, and Samsung would replace it with ENN. The back end corresponds to other frameworks such as OpenVINO for laptops and similar large mobile devices. 

\subsection{Execution Scenarios}
\label{Execution Scenarios}

The LoadGen provides two execution modes: \textit{single stream} and \textit{offline}. They reflect the  operating behavior of many mobile applications in the real world. In the single-stream scenario, the application sends a lone inference query to the SUT with a sample size of one. That size is typical of smartphones and other interactive devices where, for example, the user takes a picture and expects a timely response. The LoadGen injects a query into the SUT and waits for its completion. It then records the inference run length and sends the next query. This process repeats until the LoadGen has issued all the samples (1,024) in the task’s corresponding data set or a minimum run time of 60 seconds has passed.

In the offline scenario, the LoadGen sends all the samples to the SUT in one burst. Although the query sample size remains one, as in the single-stream scenario, the number of samples in the query is much larger. Offline mode issues 24,576 samples---enough for sufficient run time. This choice reflects applications that require multi-image processing, simultaneous processing of batched input, or concurrent application of models such as image classification and person detection to photos in an album. The implementation is usually a batched query with a batch size larger than one.

These metrics are based on best practices derived from industry feedback and open (sometimes grudging) consensus. 

\subsection{System Under Test}
\label{sec:sut}

%Smartphones and laptops can use the mobile-benchmark suite (Figure~\ref{fig:code path}). 
%The LoadGen measures inference performance at the application layer to reflect latencies that mobile-device users observe and to give developers a reference for expected user-app latencies.

We designed the LoadGen to take advantage of any mobile device type (laptop, tablets, or smartphones). Smartphones can use the reference MLPerf Android app that supports TFLite delegates and NNAPI delegates. The app queries the LoadGen, which then queries input samples for the task, loads them to memory, and tracks the time required to execute the task. For laptops, submitters can build a native command-line application. The LoadGen integrates this application, and it supports back ends such as the OpenVINO run time. The application generates logs consistent with MLPerf rules, validated by the submission checker. The number of samples necessary for performance mode and for accuracy mode remains identical to the number in the smartphone scenario. The only difference is the absence of a graphical user interface for the laptop devices.

%Companies that use proprietary vendor backends or delegates implement their back-end interface to the reference MLPerf app (Figure~\ref{fig:code path}, Section~\ref{sec:backend}). Such back ends query the correct library (TensorFlow, TFLite, the Exynos Neural Network SDK, or the SNPE SDK) to run the models.

\section{Model Optimizations}
\label{Result Submission}

We describe the process to produce high-performance results for submission. In practice, MLPerf Mobile is a competition to produce the most competitive system performance.

% \begin{figure}[t!]
%         \includegraphics[width=0.95\columnwidth]{fig/Load_generator.pdf}
%         \centering
%         \caption{Load Generator (``LoadGen'') testing the SUT.}
%         \label{fig:Load generator}
% \end{figure}

\subsection{Numerics}
\label{Submission Process}

A submitter may implement minimal changes to the model if they are mathematically equivalent or approved approximations to make the model compatible with their hardware. The rules prohibit altering the AI models to reduce their computational complexity; banned techniques include channel pruning, filter pruning, and weight skipping. However, some amount of quantization techniques are permissible.

The reference models are frozen TensorFlow FP32 checkpoints, and valid submissions must begin from these frozen graphs. Submitters can export a reference FP32 TFLite model. They can generate fixed-point models with INT8 precision from the reference FP32 models using post-training quantization (PTQ), but they cannot perform quantization-aware training (QAT). Network retraining alters the neural-network architecture, so model equivalence is difficult to verify. Also, retraining allows the submitters to use their training capabilities (e.g., neural architecture search) to boost inference throughput, changing the benchmark's nature.

Depending on submitter needs, however, we provide QAT versions of the model. All participants mutually agree on these QAT models as being comparable to the PTQ models. In general, QAT reduces accuracy loss relative to PTQ. Therefore, we chose the minimum-accuracy thresholds (``Quality Target'' in Table~\ref{tab:task_models}) on the basis of what is achievable through post-training quantization without any training data. For some benchmarks, we generated a reference INT8 QAT model using the TensorFlow quantization tools; submitters can employ it directly in the benchmark. 

Some hardware cannot deploy TensorFlow-quantized models directly, and organizations may need different fixed-point formats to match their hardware. In such cases, we only allow post-training quantization without training data from a reference model. For each model, we specify a calibration data set (typically 500 samples or images from the training or validation data set) for calibration in the PTQ process. Submitters can only use the approved calibration data set.

\subsection{Vendor Optimized Backends / SDKs}
\label{sec:backend}

\begin{figure}[t!]
\vspace{-10pt}
        \includegraphics[trim=70 50 50 70, clip, width=.9\columnwidth]{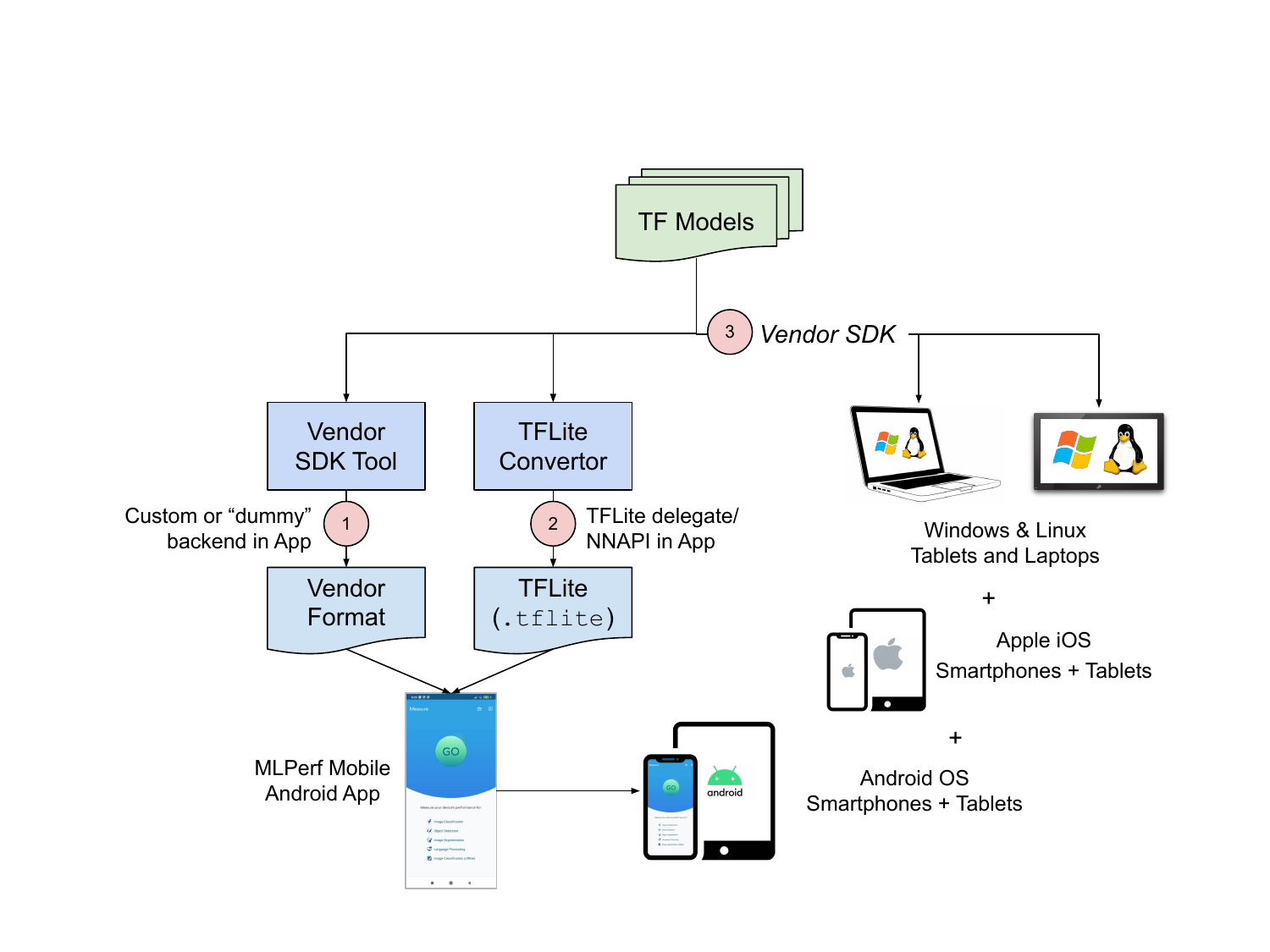}
        \centering
        \caption{Benchmark supports many devices and code paths.}
        \label{fig:code path}
\vspace{-15pt}
\end{figure}

Most chipset organizations rely on optimized software backends (or libraries) to extract the best performance from their SoCs, which often support many different hardware accelerators (CPUs, GPUs, NPUs, DSPs, etc.). Smartphone companies that use proprietary vendor backends or delegates implement their back-end interface to the reference MLPerf app (Figure~\ref{fig:code path}, Section~\ref{sec:backend}). To run the neural-network models, such back ends query the correct library (TensorFlow, TFLite, the Exynos Neural Network SDK, or the SNPE SDK). For laptops, submitters can build a native command-line application. The LoadGen integrates this application, and it supports back ends such as the OpenVINO run time. The application generates logs consistent with MLPerf rules, validated by the submission checker. The number of samples necessary for performance mode and for accuracy mode remains identical to the number in the smartphone scenario. The only difference is the absence of a graphical user interface for laptop-based devices.

Figure~\ref{fig:code path} shows how we support all this flexibility. The reference TensorFlow models are at the root of the entire process, which follows one of three paths. {Code~path~\circled{1}} allows submitters to optimize the reference TensorFlow models for implementation through a proprietary back end (e.g., SNPE for Qualcomm or ENN for Samsung), then schedule and deploy the networks on the hardware. {Code~path~\circled{2}} allows submitters to convert the reference TensorFlow models to a mobile-friendly format using an exporter. These models are then easy to deploy on the device, along with appropriate quantizations, using the TFLite delegates to access the AI-processing hardware. {Code~path~\circled{3}} allows non-smartphone submitters to run the reference TensorFlow models through non-mobile back ends (e.g., OpenVINO) on laptops and tablets with operating systems such as Windows and Linux. 

%%%%%%%%%%%%%%%%%%%%%%%%%%%
\section{Run Rules}

We developed a strict set of run rules that allow us to reproduce submitted results through an independent third party. All of the MLPerf Mobile run rules and conditions are based on best practices derived from industry input and feedback. 

\subsection{Reproducibility Guidelines}
\label{Run Rules}

\textbf{Test control.} The mobile app runs the models in a specific order. For each one, the model runs on the validation set to calculate the accuracy. Performance mode follows. Single-stream mode measures the 90th-percentile latency over at least 1,024 samples for a minimum run time of 60 seconds to achieve a stable performance result. Offline mode reports the average throughput necessary to process 24,576 samples; in current systems, the run time exceeds 60 seconds. These values were based on input from the member organizations.

\textbf{Thermal throttling.} ML models are computationally heavy and they can trigger run-time thermal throttling to cool the SoC. We recommend maintaining an air gap with proper ventilation and avoid flush contact with any surfaces. Also, we require room-temperature between 20 and \SI{25}{\celsius}. 

\textbf{Cooldown interval.} The app provides a break setting of 0--5 minutes between the individual tests to allow the phone to reach its cooldown state before starting each one. If the suite is to run multiple times, we recommend a 10-min break.

\textbf{Battery power.} The benchmark runs while the phone is battery powered, but we recommend a full charge beforehand to avoid entering power-saving mode.

% The above rules are generally inapplicable to laptops because these devices have sufficient power and cooling.

\subsection{Result Validation}
\label{Result Validation}

We require that the SUT be commercially available before publication, enabling a more tightly controlled validation, review, and audit process. The SUT includes both the hardware and software components. Submissions include all of the mobile benchmark app's log files, unedited. Post submission, all of the results are independently audited, along with any modified models and code used in the respective submissions. The vendor backend (but not the toolchain) is included. We also evaluate any private vendor SDKs to allow auditing of the model conversion process. The audit process comprises an examination of log files, models, and code for compliance with the submission rules and verification of their validity. It includes verification of the system's reported accuracy and latencies. To verify results, we build the vendor-specific app, install it on the device (in the factory-reset state), and reproduce the latency and/or throughput numbers, along with accuracy. The results are valid if our numbers are within 5\% of the submitted scores. 

\section{Insights from Benchmark Results}
\label{Performance Evaluation}

\begin{figure}[t!]
    \centering
    \includegraphics[trim=30 30 70 30, clip, width=\columnwidth]{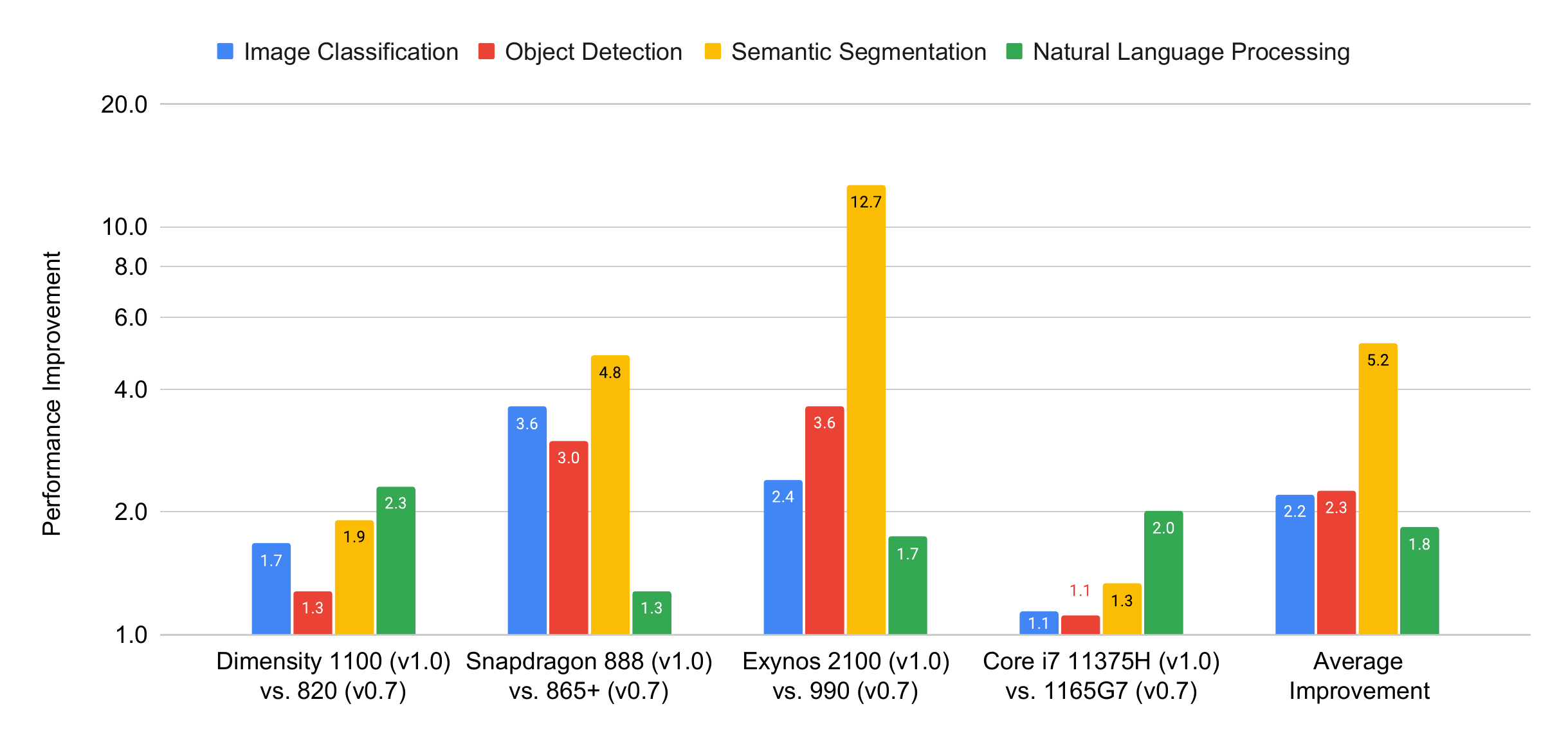}
    \caption{We see a 2$\times$ improvement between v0.7 and v1.0.}
    \label{fig:gen_latency}
\vspace{-15pt}
\end{figure}

%The benchmark was put to the test through two rounds of mobile submissions. 
To understand what is to be gained from benchmarking, we dissect the first two rounds of MLPerf Mobile submissions. We compare version 0.7 of the results~\cite{v07} to version 1.0~\cite{v10}. SoC manufacturers submitted the results using the app (see Appendix~\ref{sec:app:app}). In addition to the insights we present here, there are various other use cases of the benchmark (see Appendix~\ref{Value}).

%We analyze the results to extract interesting insights to guide future research and innovation.

%This section assesses how the benchmark performed, as in whether the benchmark meets the expectations for transparency and faithfulness and whether it reflects mobile AI hardware and software diversity. The benchmark's goal is also to challenge submitters to optimize the reference implementations we provide to maximize their performance on a mobile device under the conditions set forth by the benchmark run rules. Thereby, we ultimately answer the crucial question of whether the performance measurement benchmark leads to generational performance improvements?

\begin{figure*}[t!]
\vspace{-1em}
\centering
        \subfloat[Image classification]{\includegraphics[width=0.25\textwidth]{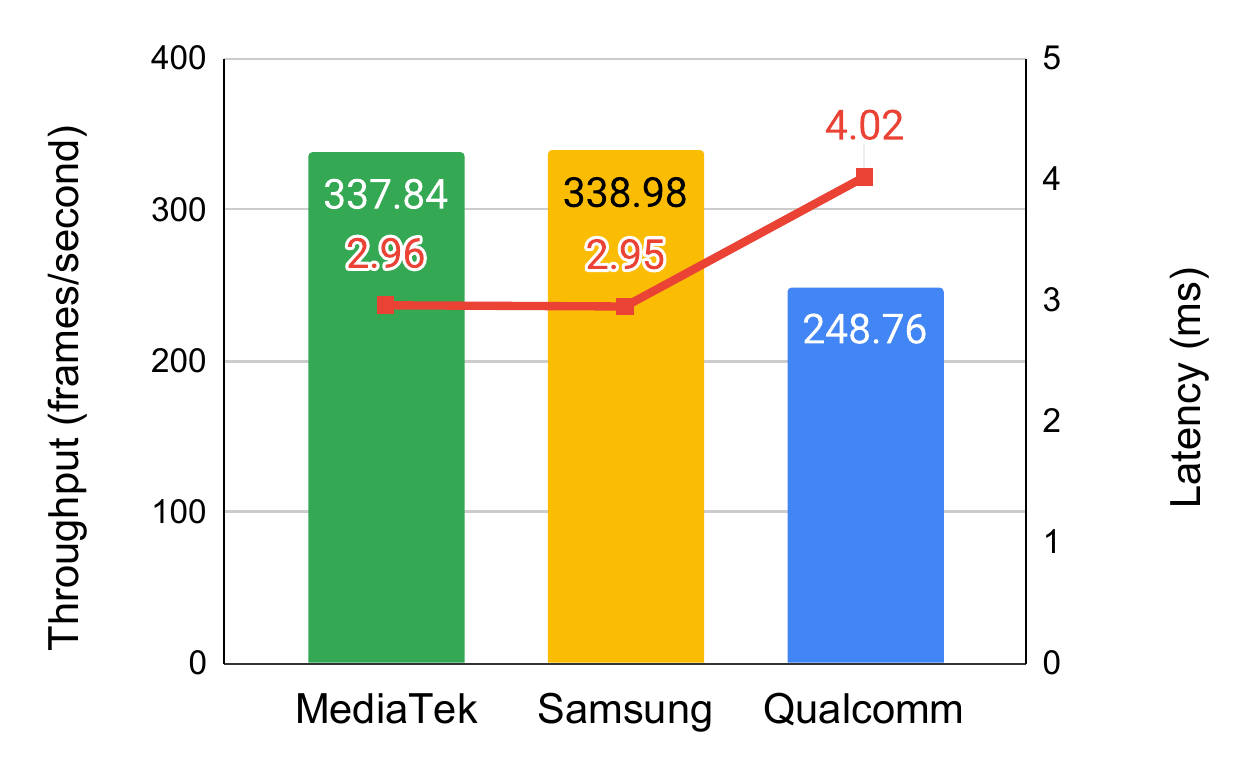}}
        \subfloat[Object detection]{\includegraphics[width=0.25\textwidth]{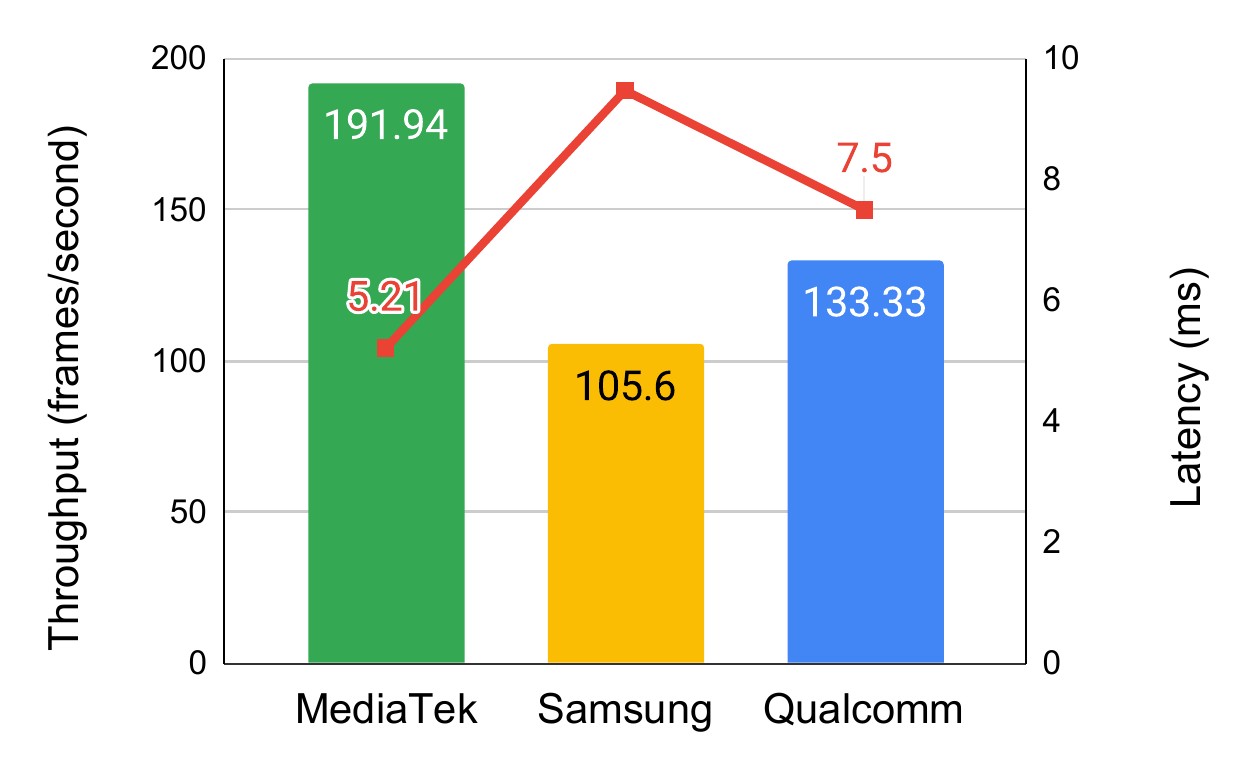}}
        \subfloat[Semantic segmentation]{\includegraphics[width=0.25\textwidth]{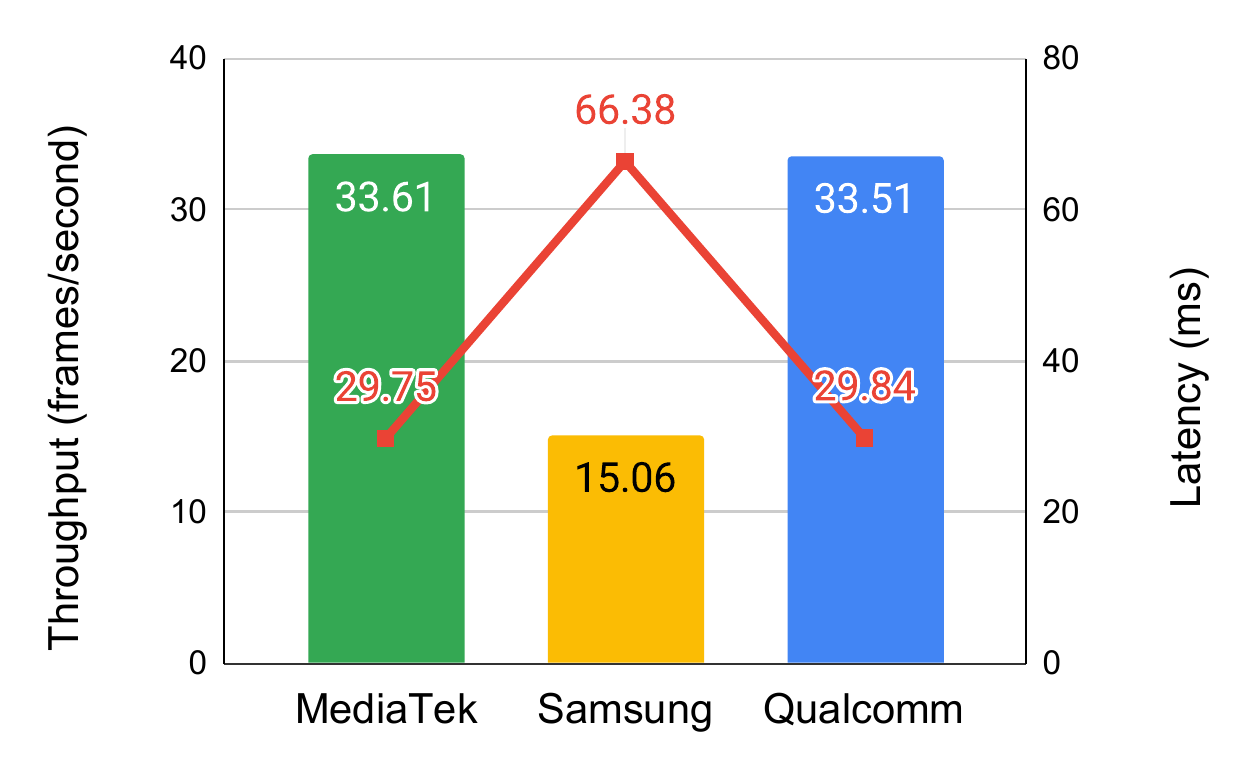}}
        \subfloat[Natural-language processing]{\includegraphics[width=0.25\textwidth]{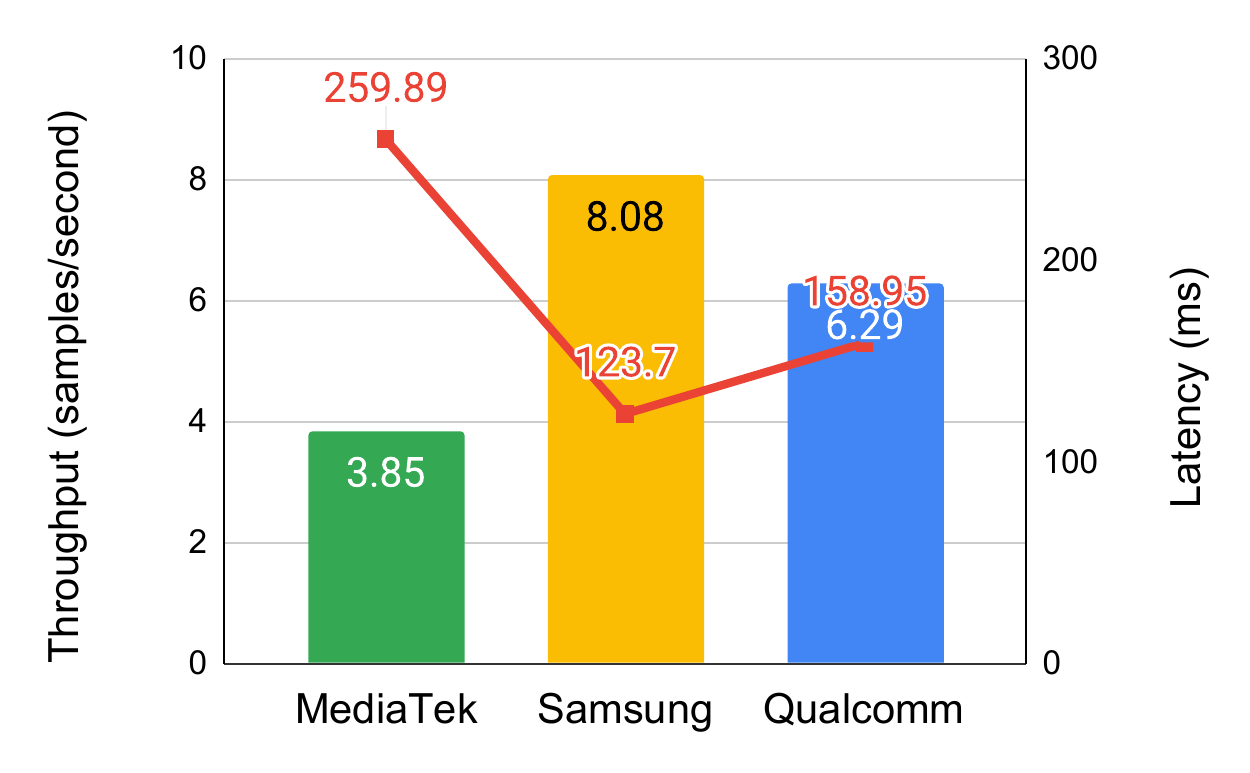}}
        \caption{Results from the first round (v0.7)~\cite{v07}. We observe similar trends in v1.0 submission round.}
        \label{fig:Performance results}
\vspace{-1em}
\end{figure*}

\subsection{Insight 1: Benchmarking Leads to Improvements}
\label{sec:measure}

Over about six months, the submitting organizations had new offerings with improved ML capabilities, both in terms of hardware and software. Results were collected on end-user consumer devices that incorporate the SoC chipsets. Figure~\ref{fig:gen_latency} compares the improvement in latency for each of the ML tasks across the two generations. Results are grouped by SoCs. The figure also shows the average gain for each of the ML tasks. The performance of the ML tasks improved by $\sim$2$\times$. The performance improvement resulted from all smartphones SoCs advancing to a new generation. The specific hardware improvements vary across each of the SoC families. We summarize the main reasons here and provide additional system improvement details in Appendix~\ref{sec:app:systems}.

The {Samsung Exynos} 2100 outperforms the 990 by 12.7$\times$ on the segmentation task. This is cause the hardware improved by more than 2$\times$. But the software also played a crucial role--its uplift was 6$\times$. Exynos 2100  has critical features that reduce data transfer between IP blocks, which are enabled in software through improved scheduling. The improved {Qualcomm Snapdragon} 888's new Hexagon 780 can perform 26 TOPS (73\% faster than 865+) along with a resigned DSP microarchitecture. The new {MediaTek's Dimensity} originally had a single core MediaTek Deep Learning Accelerator (MDLA), while the Dimensity 1100 has dual MDLA cores. Associated with these changes are the software drivers. The Dimensity 1100 uses the Neuron Delegate to replace the NNAPI Delegate, when it is possible, as NNAPI has synchronization overheard due to the intermediate hardware abstraction layer (Figure~\ref{fig:app_dev_option_b}), and also discussed in Table~\ref{tab:nnapi}. 

On the laptop front, improvements primarily came from software enhancements and minimal hardware changes. In terms of CPU frequency, {Intel's} Core i7-11375H~\cite{wilocove} is 1.1$\times$ better than i7-1165G7~\cite{xelp}; in terms of GPU frequency, i7-11375H is about 1.04$\times$ better compared to i7-1165G7. For image classification and object detection, the benchmark runs on CPU. Hence, the improvements are from an increase in CPU frequency. Segmentation and NLP models need more TOPs compared to classification and detection. For this reason, Segmentation and NLP models work best on an  integrated GPU (iGPU). Though there is a slight improvement in iGPU performance (4\% improvement), we see a large improvement in NLP performance and a marginal increase in segmentation performance. NLP improvement is due to the OpenVINO quantized kernel.

% % The submitted systems include premier 5G smartphones and high-end mobile SoCs from MediaTek, Qualcomm, and Samsung. The MediaTek chipset is a Dimensity 820 \cite{dim820} in the Xiaomi Redmi 10X smartphone; it contains MediaTek’s AI processing unit (APU) 3.0. The APU uniquely supports FP16 and INT16~\cite{lin20207}. The Qualcomm chipset is a Snapdragon 865+ \cite{sn865} in the Asus ROG Phone 3. It integrates Qualcomm's Hexagon 698 DSP, which consists of two engines that can handle AI processing exclusively. The first engine implements the Hexagon Vector Extensions (HVX), which are designed for advanced imaging and computer-vision tasks intended to run on the DSP instead of the CPU. The second, the company's AI-processor (AIP) cluster, supports the Hexagon Tensor Accelerator (HTA), which can also perform AI tasks. These engines can serve together for maximum performance or serve in isolation (depending on the compiler optimizations). The Samsung chipset is an Exynos 990 \cite{samsung} in the company's Galaxy Note 20 Ultra, which has a dual-core custom neural processing unit (NPU) to handle AI workloads. In the laptop category, Intel submitted results for its new Willow Cove CPU \cite{wilocove} and first-generation integrated Xe-LP GPU, which served as the AI accelerator \cite{xelp}. These systems collectively reflect state of the art in AI processors
% %Since the comparisons are on SingleStream, the OpenVINO compiler graph can run on a single IP block (CPU or integrated Gfx). 

\subsection{Insight 2: No One Size Fits All Tasks \& Models}

%% show that there is no approach that rules them all. 
No one solution dominates all the benchmarks at the overall benchmark-level and task-specific level. Figure~\ref{fig:Performance results} plots the single-stream results for the three smartphone chipsets on each benchmark task from the v0.7 version; the same general trend holds true for the v1.0 version. 
The figure shows throughput and latency results. Each chipset offers a unique differentiable value. MediaTek’s Dimensity scored the highest in object-detection and image-segmentation throughput. Samsung's Exynos performed well on image classification and NLP, where it achieved the highest scores. Qualcomm's Snapdragon is competitive for image segmentation and NLP. 

The image-classification task also employs an offline mode for batch processing; here, Exynos delivered 674.4 frames per second (FPS) and Snapdragon delivered 605.37 FPS. These data points are not shown in Figure~\ref{fig:Performance results}. Also, not all submitters are required to submit to this offline scenario.

Table~\ref{tab:submission_table} shows the details behind Figure~\ref{fig:Performance results}. Because MLPerf Mobile emphasizes transparency, the table includes specifics for how a benchmark is executed in the single-stream mode and the offline mode. The table shows the numerics (top of cell), framework (middle of cell), and accelerator (bottom of cell) used to produce the results. The table shows that various hardware combinations are used to achieve good mobile AI performance. No one hardware unit dominates all ML tasks. In all cases, the CPU is the backbone (not included in the table) that is orchestrating the overall execution---including doing pre- and post-processing and other tasks the benchmark does not measure. In contrast, as shown in the last row of each cell, the GPU, DSPs, NPUs, and AIPs are focused on delivering high-performance AI execution. 

\begin{table*}[t!]
\resizebox{\linewidth}{!}{
\begin{tabular}{clllll}
\hline
\multicolumn{1}{|l|}{}                  & \multicolumn{1}{c|}{\begin{tabular}[c]{@{}c@{}}\textbf{Image Classification}\\ \textbf{(single-stream)}\end{tabular}} & \multicolumn{1}{c|}{\begin{tabular}[c]{@{}c@{}}\textbf{Image Classification}\\ \textbf{(offline)}\end{tabular}} & \multicolumn{1}{c|}{\begin{tabular}[c]{@{}c@{}}\textbf{Object Detection}\\ \textbf{(single-stream)}\end{tabular}} & \multicolumn{1}{c|}{\begin{tabular}[c]{@{}c@{}}\textbf{Image Segmentation}\\ \textbf{(single-stream)}\end{tabular}} & \multicolumn{1}{c|}{\begin{tabular}[c]{@{}c@{}}\textbf{Natural-Language Processing}\\ \textbf{(single-stream)}\end{tabular}} \\ \hline
\multicolumn{1}{|c|}{}                  & \multicolumn{1}{c|}{\textit{ImageNet}}                                                              & \multicolumn{1}{c|}{\textit{ImageNet}}                                                        & \multicolumn{1}{c|}{\textit{COCO}}                                                              & \multicolumn{1}{c|}{\textit{ADE20K}}                                                              & \multicolumn{1}{c|}{\textit{Squad}}                                                                        \\ \hline
\multicolumn{1}{|c|}{}                  & \multicolumn{1}{c|}{\textit{MobileNetEdge}}                                                         & \multicolumn{1}{c|}{\textit{MobileNetEdge}}                                                   & \multicolumn{1}{c|}{\textit{SSD-MobileNet v2}}                                                 & \multicolumn{1}{c|}{\textit{DeepLab v3+ - MobileNet v2}}                                            & \multicolumn{1}{c|}{\textit{MobileBERT}}                                                                   \\ \hline\hline
\multicolumn{1}{|c|}{\begin{tabular}[c]{@{}c@{}}\textit{MediaTek}\\ \textit{Dimensity 820}\\ \textit{(smartphone)}\end{tabular}} & \multicolumn{1}{l|}{\begin{tabular}[c]{@{}l@{}}UINT8,\\ NNAPI (neuron-ann),\\ APU\end{tabular}}     & \multicolumn{1}{l|}{Not applicable}                                                                      & \multicolumn{1}{l|}{\begin{tabular}[c]{@{}l@{}}UINT8,\\ NNAPI (neuron-ann),\\ APU\end{tabular}} & \multicolumn{1}{l|}{\begin{tabular}[c]{@{}l@{}}UINT8,\\ NNAPI (neuron-ann),\\ APU\end{tabular}}   & \multicolumn{1}{l|}{\begin{tabular}[c]{@{}l@{}}FP16,\\ TFLite delegate,\\ Mali-GPU\end{tabular}}           \\ \hline
\multicolumn{1}{|c|}{\begin{tabular}[c]{@{}c@{}}\textit{Samsung}\\ \textit{Exynos 990}\\ \textit{(smartphone)}\end{tabular}}  & \multicolumn{1}{l|}{\begin{tabular}[c]{@{}l@{}}INT8,\\ ENN,\\ NPU+CPU\end{tabular}}                     & \multicolumn{1}{l|}{\begin{tabular}[c]{@{}l@{}}INT8,\\ ENN,\\ NPU+CPU\end{tabular}}               & \multicolumn{1}{l|}{\begin{tabular}[c]{@{}l@{}}INT8,\\ ENN,\\ NPU+CPU\end{tabular}}          & \multicolumn{1}{l|}{\begin{tabular}[c]{@{}l@{}}INT8,\\ ENN,\\ NPU+GPU\end{tabular}}            & \multicolumn{1}{l|}{\begin{tabular}[c]{@{}l@{}}FP16,\\ ENN,\\ GPU\end{tabular}}                            \\ \hline
\multicolumn{1}{|c|}{\begin{tabular}[c]{@{}c@{}}\textit{Qualcomm}\\ \textit{Snapdragon 865+}\\ \textit{(smartphone)}\end{tabular}} & \multicolumn{1}{l|}{\begin{tabular}[c]{@{}l@{}}UINT8,\\ SNPE,\\ HTA\end{tabular}}                   & \multicolumn{1}{l|}{\begin{tabular}[c]{@{}l@{}}UINT8,\\ SNPE,\\ AIP (HTA+HVX)\end{tabular}}             & \multicolumn{1}{l|}{\begin{tabular}[c]{@{}l@{}}UINT8,\\ SNPE,\\ HTA\end{tabular}}               & \multicolumn{1}{l|}{\begin{tabular}[c]{@{}l@{}}UINT8,\\ SNPE,\\ HTA\end{tabular}}                 & \multicolumn{1}{l|}{\begin{tabular}[c]{@{}l@{}}FP16,\\ TFLite delegate,\\ GPU\end{tabular}}             \\ \hline\hline
\multicolumn{1}{|c|}{\begin{tabular}[c]{@{}c@{}}\textit{Intel}\\ \textit{Core i7-1165G7}\\ \textit{(laptop)}\end{tabular}}    & \multicolumn{1}{l|}{\begin{tabular}[c]{@{}l@{}}INT8,\\ OpenVINO,\\ CPU\end{tabular}}                & \multicolumn{1}{l|}{\begin{tabular}[c]{@{}l@{}}INT8,\\ OpenVINO,\\ CPU+GPU\end{tabular}}      & \multicolumn{1}{l|}{\begin{tabular}[c]{@{}l@{}}INT8,\\ OpenVINO,\\ CPU\end{tabular}}            & \multicolumn{1}{l|}{\begin{tabular}[c]{@{}l@{}}INT8,\\ OpenVINO,\\ GPU\end{tabular}}              & \multicolumn{1}{l|}{\begin{tabular}[c]{@{}l@{}}INT8,\\ OpenVINO,\\ GPU\end{tabular}}                       \\ \hline
\multicolumn{1}{l}{}                    &                                                                                                     &                                                                                               &                                                                                                 &                                                                                                   &
\end{tabular}
}
\vspace{-12pt}
\caption{Myriad combinations of numerics, software run times, and hardware, reinforcing the need for transparency.}
\label{tab:submission_table}
\end{table*}

\subsection{Insight 3: Accelerator Level Parallelism is Here}

The results generally highlight another important point that to deliver the best performance vendors rely on multiple accelerators concurrently, more recently referred to as accelerator level parallelism (ALP)~\cite{hill2019accelerator}. Relying on multiple accelerators allows the ML compiler framework to intelligently schedule portions of the neural network graph operations to different execution engines that best match the operation's needs. Table~\ref{tab:submission_table} includes details for the image classification offline mode. In offline mode, the focal point is the throughput of the system (not latency). 

Across the offline mode results (second column) in Table~\ref{tab:submission_table}, multiple accelerators are used concurrently to maximize performance. For example, Exynos uses both the NPU and the CPU together. Similarly, on the Snapdragon chipsets, the Hexagon Tensor Accelerator (HTA) and Hexagon Vector Extensions (HVX) are used concurrently as part of the AI processing (AIP) cluster. On the core~i7 SoC, both the CPU and the integrated GPU are used simultaneously. It is uncommon to exercise ALP in the latency-bounded single-stream scenario because the overhead of managing multiple concurrent accelerators can quickly become a bottleneck.

% The submission results highlight an important point: they reflect the variety of hardware and software combinations we discussed earlier (Section~\ref{Benchmarking Challenges}). All mobile SoCs rely on a generic processor, but the AI-performance results were from AI accelerators using different software frameworks. Transparency into how the results were generated is crucial. 

% Figure~\ref{fig:codepathway} shows the potential code paths for producing the submission results. The dashed lines represent mere possibilities, whereas the solid lines indicate actual submissions. Looking only at Figure~\ref{fig:codepathway} is insufficient to determine which paths produce high-quality results. Other code paths would have yielded a different performance result. Therefore, benchmark-performance transparency is essential: it reveals which code paths were taken, making the performance results reproducible and informative for consumers. 

% Exposing each of these details is important because the many execution paths in Figure~\ref{fig:codepathway} can drastically affect a device's performance. 

\subsection{Insight 4: ML Frameworks Play a Crucial Role}

Given the heterogeneity of the  ecosystem, the NNAPI runtime module is a library sitting between an app and its backend drivers. It is designed to be a common baseline for ML on Android devices and to distribute that workload across ML-processor units, such as CPUs, GPUs, DSPs, and NPUs. But nearly all submissions in Table~\ref{tab:submission_table} make use of proprietary frameworks (Figure~\ref{fig:code path}). Vendor SDKs, such as ENN and SNPE, give SoC vendors control over their SoC's performance. They can control which processor unit or accelerator to use for AI and what optimizations to apply.

All laptop submissions employ INT8 and achieve the desired accuracy on vision and language models. For single-stream mode, because just one sample is available per query, some models are incapable of fully utilizing the GPU's computational resources. Therefore, the back end must choose between the CPU and GPU to deliver the best overall performance. For example, small models such as MobileNetEdgeTPU use the CPU. For offline mode, multiple samples are available as a single query, so inference employs both the CPU and GPU. This level of detailed control is strongly tied to the system's capabilities and only the hardware vendors are well-suited for this sort of fine-grained optimization. 

\begin{table}[h!]
\centering
\resizebox{\columnwidth}{!}{
\begin{tabular}{|l|l|l|l|}
\hline
\multicolumn{1}{|c|}{\textbf{\begin{tabular}[c]{@{}c@{}}\it MediaTek\\ \it Dimensity 1100\end{tabular}}} & \multicolumn{1}{c|}{\textbf{\begin{tabular}[c]{@{}c@{}}Image\\ Classification\end{tabular}}} & \multicolumn{1}{c|}{\textbf{\begin{tabular}[c]{@{}c@{}}Object\\ Detection\end{tabular}}} & \multicolumn{1}{c|}{\textbf{\begin{tabular}[c]{@{}c@{}}Image\\ Segmentation\end{tabular}}} \\ \hline
NNAPI Delegate         & 2.48 ms                                                                             & 5.05 ms                                                                         & 20.56 ms                                                                          \\ \hline
Neuron Delegate        & 2.23 ms                                                                             & 4.77 ms                                                                         & 20.02 ms                                                                          \\ \hline\hline
\% Improvement          & 10.08\%                                                                                & 5.54\%                                                                            & 2.70\%                                                                              \\ \hline
\end{tabular}
}
\caption{Vendor-optimized delegates tend to perform better.}
\label{tab:nnapi}
\vspace{-1em}
\end{table}

Table~\ref{tab:nnapi} shows the performance difference for v1.0 tasks between using the generic NNAPI delegate and the optimized Neuron Delegate. The latter has full support for MediaTek's Dimensity 1100. Using the optimized framework driver can lead to over 10\% difference in performance. This is because the standard NNAPI driver does not yet fully support multi-MDLA, which we discussed previously in Section~\ref{sec:measure}.

\subsection{Insight 5: Numerics Still Matter for Some Tasks}
\label{Execution Diversity}

Quantization is crucial for mobile deployments because quantized inference runs faster and provides better performance and memory bandwidth than FP32 \cite{han2015deep}. The accuracy tradeoff for quantized models (especially since no retraining is allowed in the benchmark) is tolerable in smartphones, which seldom perform safety-critical tasks. 

However, it is not always about INT8. Mobile-device designers prefer both INT8 and FP16 (Table~\ref{tab:submission_table}). All the mobile-vision tasks employ INT8 heavily. Most vendors rely on this format because it enables greater performance and consumes less power, preserving device battery life. NLP favors FP16, which requires more power than INT8 but offers better accuracy. More importantly, submitters use FP16 because most AI engines today lack efficient support for non vision tasks. 

\section{Related Work}
\label{Related work}

Compared to existing mobile AI benchmarks, MLPerf Mobile is different in the following ways, all of which we believe are \textit{requirements} to drive the industry forward:

\vspace{-1em}
\begin{enumerate}
  \setlength\itemsep{0em}
    \item need \textbf{a system-level machine learning benchmark, designed to exercise the complete mobile system}, compared to micro-benchmarking small pieces.
    \item put \textbf{accuracy first and measure performance with respect to a minimum quality target}, based on community consensus, rather than measuring performance for unchecked and arbitrary model quality targets.
    \item be an \textbf{open-source benchmark that provides transparency into results and implementations} where submitters are required to submit their code for audits
    \item  \textbf{support many vendor backends/SDKs} and NNAPI and TFLite delegates to unleash the SoCs' capabilities.
    \item be \textbf{driven and audited by the industry,} which fosters fair and representative performance evaluations.
\end{enumerate}

Due to space constraints and the lack of transparency into existing benchmarks, we indicate at least a few major requirements that are \textit{missing} from prior art in Table~\ref{tab:compare}. We provide additional details to the extent possible in Appendix~\ref{sec:app:related}.

\begin{table}[t]
\centering
\small
\resizebox{\columnwidth}{!}{
\begin{tabular}{|l|c|c|c|c|c|}
\hline
              & \multicolumn{1}{c|}{\bf Req. 1} & \multicolumn{1}{c|}{\bf Req. 2} & \multicolumn{1}{c|}{\bf Req. 3} & \multicolumn{1}{c|}{\bf Req. 4} & \multicolumn{1}{c|}{\bf Req. 5} \\\hline
Aitutu        &  \gmark & \rmark                      &  \rmark               &   \gmark                    &  \rmark                    \\\hline
AI-Benchmark  &  \gmark & \rmark                      &  \rmark                     &  \rmark                    &  \rmark                    \\\hline
AIMark        & \gmark  & \rmark                      &  \rmark               &  \gmark                     &  \rmark                    \\\hline
Android MLTS  & \rmark  & \rmark                      &  \gmark            &   \gmark              &  \rmark                    \\\hline
GeekBenchML   & \gmark  &  \rmark                    &  \rmark                & \rmark                     &  \rmark                    \\\hline
Neural Scope  & \gmark  &  \rmark                     &  \rmark               &  \rmark                    &  \rmark                    \\\hline
TF Lite       &  \rmark& \rmark                      &   \gmark               &  \gmark                     &  \rmark                    \\\hline
UL Procyon AI & \gmark  & \rmark                      & \rmark                      &  \rmark                    &  \rmark                    \\\hline
Xiaomi        & \gmark  &  \rmark                     &  \gmark                     &  \rmark                     &  \rmark  \\\hline\hline
MLPerf Mobile &\gmark   & \gmark                      &  \gmark                   & \gmark                      &  \gmark                   \\\hline
\end{tabular}
}
\caption{Comparison to other mobile ML benchmarks where they are missing (\xmark) at least one (or more) requirement(s).}
\vspace{-2em}
\label{tab:compare}
\end{table}

One crucial requirement missing from all the other benchmarks is that they are \emph{not} driven by industry collaboration and consensus. A community-driven effort leads to the adoption of best practices that are fair to everyone. Moreover, it helps amortize the overhead of developing a complex benchmark like MLPerf to meet all five requirements. It is for this exact reason that Table~\ref{tab:compare} shows that the other benchmarks are each missing at least one major feature requirement. 

Another major difference is measuring performance with respect to an minimum accuracy target. Table~\ref{tab:task_models} shows the  targets we set based on academic community and industry consensus, as well as ML consumers' feedback. Our targets are all $>$93\% FP32 (Table~1), even for int8 models. Other benchmarks show results for arbitrary accuracy targets that are less (if at all) meaningful. For example, GeekBenchML reports performance for int8 models with 52\% of FP32 accuracy for object detection and 81\% of FP32 accuracy for image classification~\cite{geekbenchMLScore}. We leave it up to the reader to interpret if such results are practically useful.

\section{Conclusion}
\label{Conclusion}

% Machine-learning inference has many potential applications. Building a benchmark that encapsulates this broad spectrum is challenging. In this paper, we focused on smartphones and the mobile-PC ecosystem, which is rife with hardware and software heterogeneity. Coupled with the life-cycle complexities of mobile deployments, this heterogeneity makes benchmarking mobile-AI performance overwhelmingly difficult. To bring consensus, we developed MLPerf Mobile Inference. Many leading organizations have joined us in building a unified benchmark that meets disparate needs. The unique value of MLPerf Mobile is less in the benchmarks, rules, and metrics, but more in the value that the industry creates for itself, benefiting everyone. 

%Mobile AI performance analysis is deceptively simple; it looks simple when, in fact, it is  sophisticated. Demystifying mobile AI performance results requires us to build new instruments (tools, benchmarks, etc.) to conduct experiments more systematically and make repeatable measurements that ensure forward progress in innovation. 

%Quantitative reasoning is the cornerstone of computer architecture and ML system design. Hence, it is essential to have widely accepted benchmarks, datasets and metrics. 
This paper outlines the challenges, issues, and opportunities we faced over two years of research and development to engineer an acceptable benchmark across multiple competing organizations. We hope that the observations we make and the app we provide enables architects, designers, and developers to push the frontiers of mobile ML systems. There are several ongoing developments---expanding the suite, measuring end-to-end performance, evaluating iOS, ML framework benchmarking, measuring power and supporting rolling submissions. Details are presented in Appendix~\ref{Future work}.

\bibliography{example_paper}
\bibliographystyle{mlsys2022}

%%%%%%%%%%%%%%%%%%%%%%%%%%%%%%%%%%%%%%%%%%%%%%%%%%%%%%%%%%%%%%%%%%%%%%%%%%%%%%%
%%%%%%%%%%%%%%%%%%%%%%%%%%%%%%%%%%%%%%%%%%%%%%%%%%%%%%%%%%%%%%%%%%%%%%%%%%%%%%%
% SUPPLEMENTAL CONTENT AS APPENDIX AFTER REFERENCES
%%%%%%%%%%%%%%%%%%%%%%%%%%%%%%%%%%%%%%%%%%%%%%%%%%%%%%%%%%%%%%%%%%%%%%%%%%%%%%%
%%%%%%%%%%%%%%%%%%%%%%%%%%%%%%%%%%%%%%%%%%%%%%%%%%%%%%%%%%%%%%%%%%%%%%%%%%%%%%%
\appendix
\clearpage
\section*{\centering APPENDIX}

\addtocontents{toc}{\cftpagenumberson{chapter}}

\section{MLPerf Mobile App}
\label{sec:app:app}

Measuring mobile AI performance in a fair, reproducible, and useful manner is challenging but not intractable. 

As discussed previously in Section~\ref{Benchmarking Challenges}, the need for transparency owes to the massive hardware and software diversity, which is often tightly coupled with the intricacies of deployment scenarios, developer options, OEM lifecycles, and so on. MLPerf Mobile focuses on ensuring transparency for consumers by packaging the submitted code into an app. the app is readily accessible to the public~\cite{MLCmobile-app}. Moreover, the associated ``backend'' code is readily available for review under the MLPerf rules.

\begin{figure*}[t!]
        \centering
        \subfloat[Startup screen]{
        \includegraphics[width=0.19\linewidth]{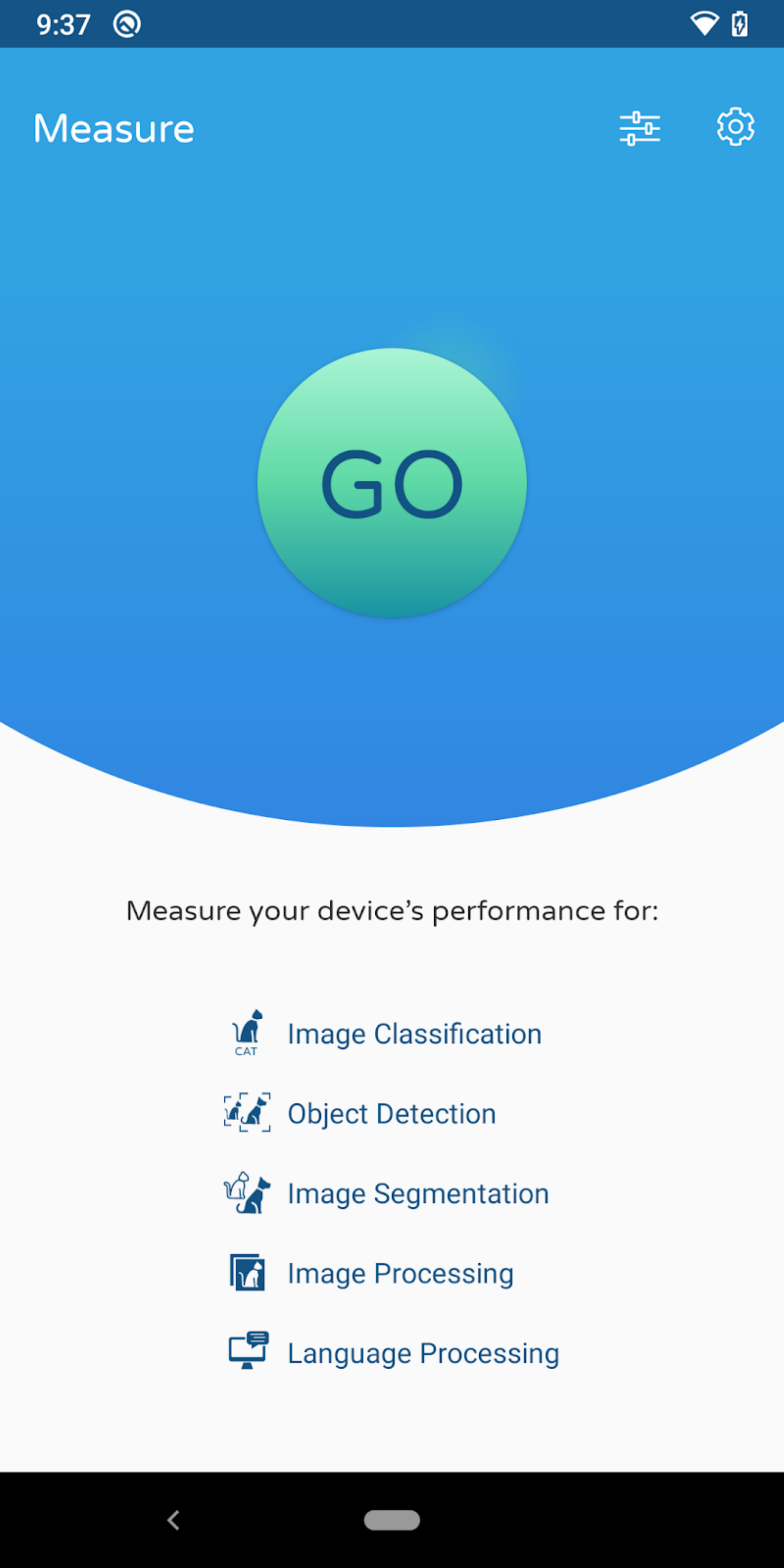}
        \label{fig:mlperfapp_startup}
        }
        \hfill
        \subfloat[Running]{\includegraphics[width=0.19\linewidth]{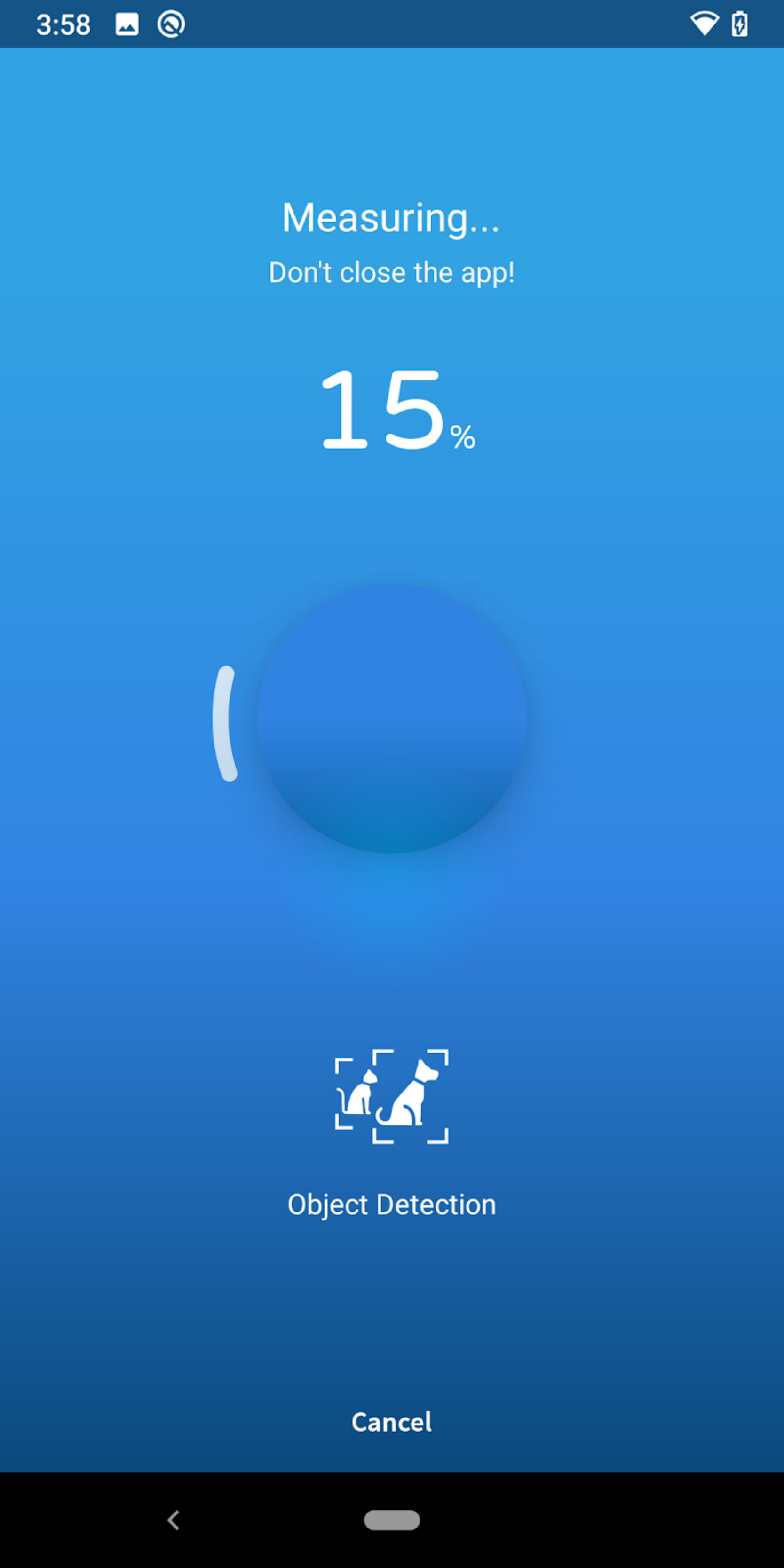}
        \label{fig:mlperfapp_go}
        }
        \hfill
        \subfloat[Reporting results]{\includegraphics[width=0.19\linewidth]{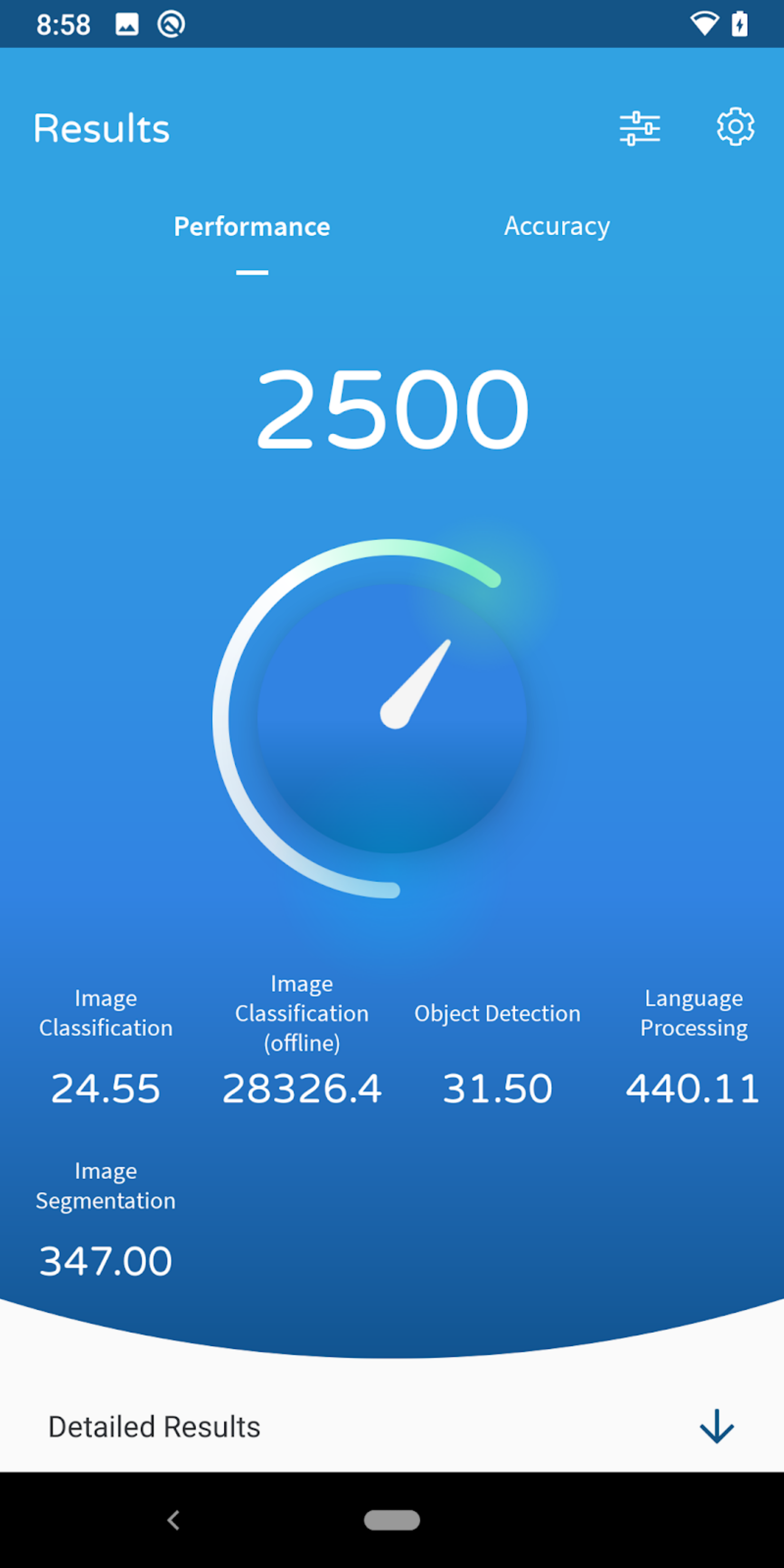}
        \label{fig:mlperfapp_run}
        }
        \hfill
        \subfloat[Run details]{\includegraphics[width=0.19\linewidth]{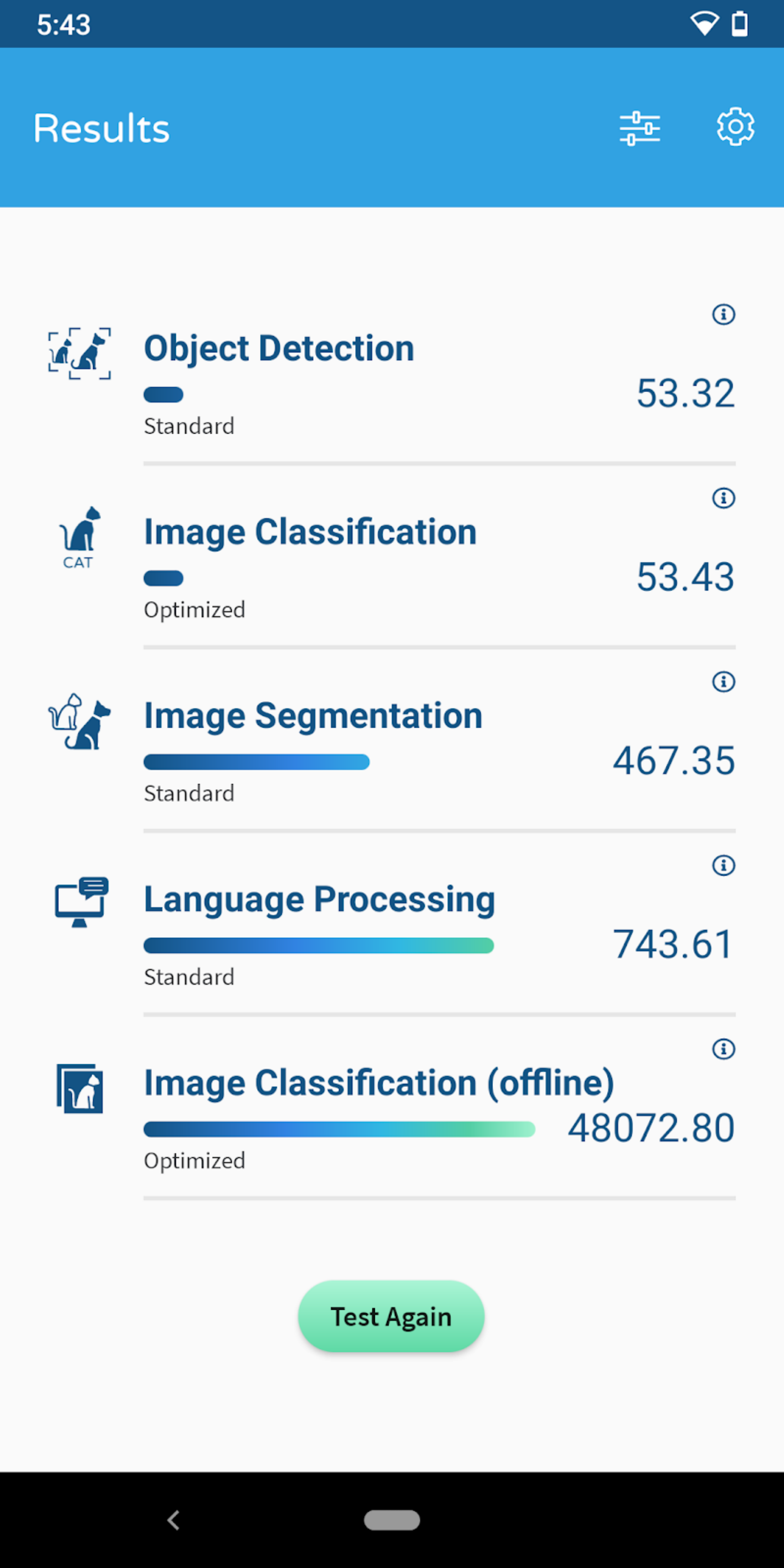}
        \label{fig:mlperfapp_results}
        }
        \hfill
        \subfloat[Configuration settings]{\includegraphics[width=0.19\linewidth]{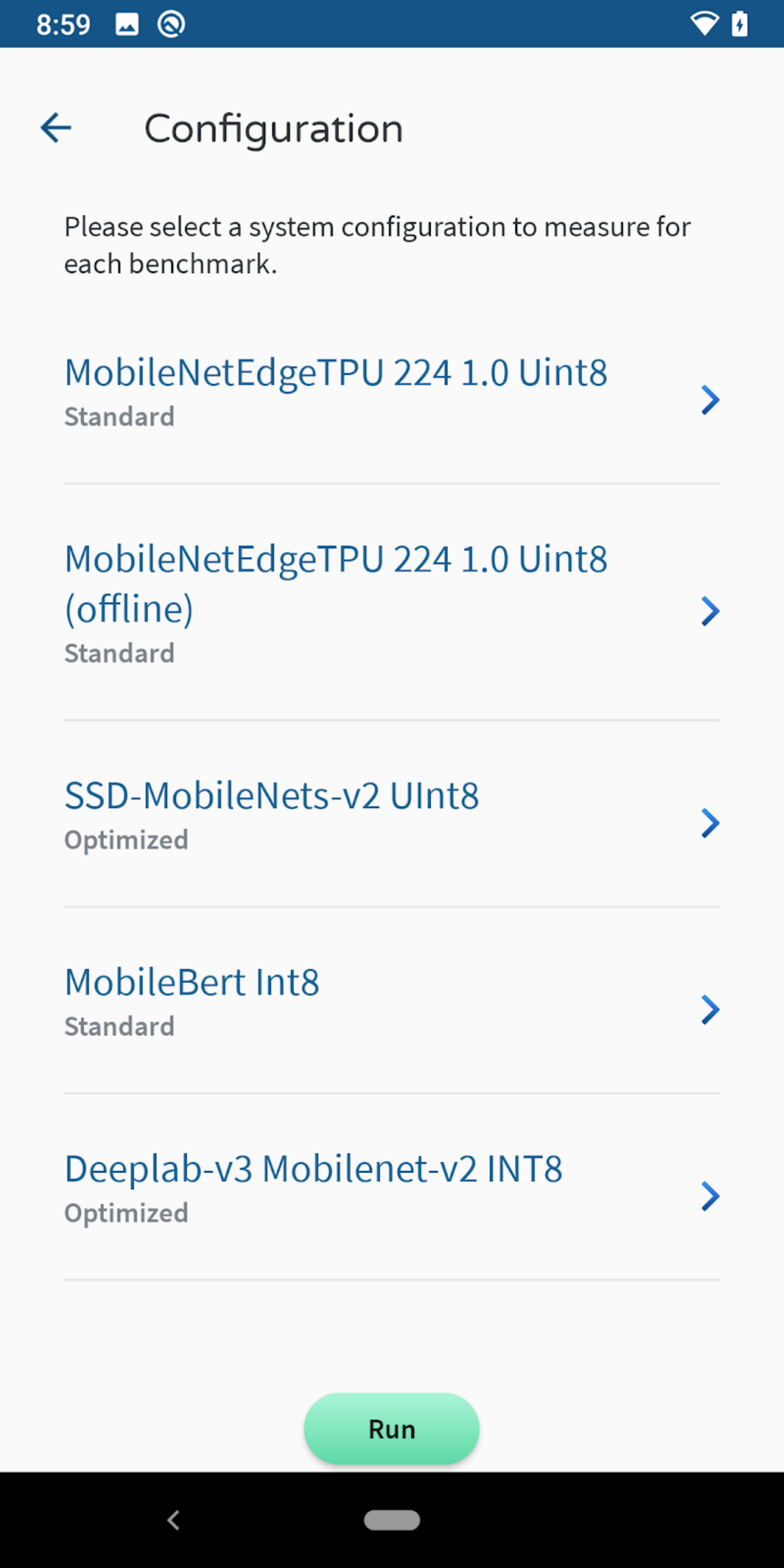}
        \label{fig:mlperfapp_config}
        }
        \caption{MLPerf Mobile app.}
        \label{fig:MLPerf Mobile app on Android.}
\end{figure*}

Figure~\ref{fig:mlperfapp_startup} shows the MLPerf Mobile startup screen.  With a simple tap on the ``Go'' button, the app runs all benchmarks by default, following the prescribed run rules (Figure~\ref{fig:mlperfapp_go}), and clearly displays the results. It reports both performance and accuracy for all benchmark tasks (Figure~\ref{fig:mlperfapp_run}) and permits the user to view results for each one (Figure~\ref{fig:mlperfapp_results}). Furthermore, the configuration that generates the results is also transparent (Figure~\ref{fig:mlperfapp_config}). This allows users to understand what hardware and software configuration yielded the high-performance results. This transparency level is the critical missing feature from many off-the-shelf benchmarks as described in the Related Work (Section~\ref{Related work}).

\section{Myriad Use Cases}
\label{Value}

While we highlight a few interesting observations in Section~\ref{Performance Evaluation}, there are many other use cases in the wild. 

\textbf{Application developers} want to know what real-world performance may look like on a device. The benchmark provides insight into the software frameworks on the various ``phones'' (i.e., SoCs) for these application developers. More specifically, the benchmark can help them quickly identify the most optimal solution for a given platform. For application developers who deploy their products ``into the wild,'' the benchmark and the various machine-learning tasks offer a deep perspective on the end-user experience for an actual mobile AI application.  

\textbf{OEMs} want to standardize the performance assessment methodology across different mobile chipset offerings. SoC vendors employ the same tasks, models, data sets, metrics, and run rules, making the results comparable and reproducible. Given the hardware ecosystem's vast heterogeneity, the standardization that the benchmark provides is vital for progress. 

\textbf{Model designers} want to package new machine learning models into the mobile app so that organizations can then easily share and reproduce the results. The app, coupled with the LoadGen, allows model designers to test and evaluate the model’s performance on a real device rather than using operation counts and model size as heuristics to estimate performance. This feature closes the gap between model designers and hardware vendors---groups that have struggled to share information efficiently and effectively. 

\textbf{Mobile users} wants to make informed purchasing decisions. Many users want to know whether upgrading their phone to the latest chipset will meaningfully improve their experience. To this end, they want public, accessible information about various devices---something MLPerf Mobile provides. In addition, some power users want to measure their device’s performance and share that information with performance-crowdsourcing platforms. Both are important reasons for having an easily reproducible mechanism for measuring mobile-AI performance.

\textbf{Researchers} often require reproducibility to push state-of-the-art technologies. As such, researchers can employ the mobile-app framework to test their methods and techniques for improving model performance, quality, or both. The framework is open-source and freely accessible to academia~\cite{MLCmobile-app}. The app can enable researchers to integrate their ML optimizations and reproduce more recent results from the literature. 

\textbf{Technical analysts} rely on reproducibility and transparency to provide an ``apples-to-apples'' comparison to assess generational performance improvements. MLPerf Mobile makes it easy to reproduce vendor-claimed results and interpret them because it shows how the device achieves a particular performance number and how it is using the hardware accelerator. 

\section{System Improvements: v0.7 to v1.0}
\label{sec:app:systems}

In this section, we provide a discussion about the hardware and system improvements that led to the significant performance improvement between benchmark v0.7 and v1.0. 

The \textbf{Samsung Exynos} Exynos 990~\cite{samsung} has a dual-core neural processing unit (NPU) and an Arm Mali-G77 GPU. The Exynos 2100 has an 8-core CPU on a tri-cluster architecture with more than 30\% improved multi-core performance, Arm Mali-G78 MP14 GPU with more than 40\% performance improvement, and an AI engine with a triple-core NPU and a DSP, based on 5nm EUV delivering powerful performance and more than 2$\times$ the efficiency than the previous generation. But the software also played a crucial role. The uplift was 6$\times$. Exynos 2100  has critical features that reduce data transfer between IP blocks, which are enabled in software through improved scheduling.

The \textbf{Qualcomm's Snapdragon} 865+~\cite{sn865} uses the Hexagon 698 processor for AI acceleration. It clocks in at 15 TOPS and has a Adreno 650 GPU. The improved Snapdragon 888's new Hexagon 780 can perform 26 TOPS (73\% faster than 865+). It also boasts a redesigned DSP microarchitecture. In the Hexagon 865 DSP, the scalar, vector and tensor execution engines were discrete independent blocks. In the Hexagon 780, the new IP block fuses all the scalar, tensor, and vector capabilities into a single monolithic IP, increasing ML workloads' performance for the second version (v1.0).

The \textbf{MediaTek's Dimensity} 820~\cite{dim820} uses an AI processing unit (APU) 3.0, an FP16 and INT16-capable accelerator optimized for camera and imaging functions~\cite{lin20207}. The SoC also has a 5-core GPU. The new Dimensity 1100 is similar in features except that it is manufactured on a 6~nm vs 7~nm process node with a more powerful GPU that is helpful for ML-task acceleration. The Dimensity 820 has single core MediaTek Deep Learning Accelerator (MDLA), while the Dimensity 1100 has dual MDLA cores. Associated with these changes are the software drivers. The Dimensity 1100 uses the Neuron Delegate to replace the NNAPI Delegate, when it is possible, as NNAPI has synchronization overheard due to the intermediate hardware abstraction layer (Figure~\ref{fig:app_dev_option_b}).

%The tri-core NPU inside the new Exynos 2100 can perform up to 26 TOPS, while the Exynos 990’s dual-core NPU is capable of 15 TOPS at its peak performance.

%On the laptop front, improvements primarily came from software enhancements and minimal hardware changes. 
For v0.7, \textbf{Intel's Willow Cove}~\cite{wilocove} includes its CPU and first-generation integrated Xe-LP GPU, a Tiger Lake i7-1165G7~\cite{xelp}. For v1.0 submission, they used a TGL i7-11375H laptop. In terms of CPU frequency, i7-11375H is about 1.1$\times$ better than i7-1165G7; in terms of GPU frequency, i7-11375H is about 1.04$\times$ better compared to i7-1165G7. For image classification and object detection, the benchmark runs on CPU. Hence, the improvements are from an increase in CPU frequency. Segmentation and NLP models need more TOPs compared to classification and detection. For this reason, Segmentation and NLP models work best on an integrated GPU (iGPU). Though there is a slight improvement in iGPU performance (4\% improvement), we see a large improvement in NLP performance and a marginal increase in segmentation performance. NLP improvement is due to the OpenVINO quantized kernel.

% The submitted systems include premier 5G smartphones and high-end mobile SoCs from MediaTek, Qualcomm, and Samsung. The MediaTek chipset is a Dimensity 820 \cite{dim820} in the Xiaomi Redmi 10X smartphone; it contains MediaTek’s AI processing unit (APU) 3.0. The APU uniquely supports FP16 and INT16~\cite{lin20207}. The Qualcomm chipset is a Snapdragon 865+ \cite{sn865} in the Asus ROG Phone 3. It integrates Qualcomm's Hexagon 698 DSP, which consists of two engines that can handle AI processing exclusively. The first engine implements the Hexagon Vector Extensions (HVX), which are designed for advanced imaging and computer-vision tasks intended to run on the DSP instead of the CPU. The second, the company's AI-processor (AIP) cluster, supports the Hexagon Tensor Accelerator (HTA), which can also perform AI tasks. These engines can serve together for maximum performance or serve in isolation (depending on the compiler optimizations). The Samsung chipset is an Exynos 990 \cite{samsung} in the company's Galaxy Note 20 Ultra, which has a dual-core custom neural processing unit (NPU) to handle AI workloads. In the laptop category, Intel submitted results for its new Willow Cove CPU \cite{wilocove} and first-generation integrated Xe-LP GPU, which served as the AI accelerator \cite{xelp}. These systems collectively reflect state of the art in AI processors
%Since the comparisons are on SingleStream, the OpenVINO compiler graph can run on a single IP block (CPU or integrated Gfx). 

\section{Detailed Prior Art Comparison}
\label{sec:app:related}
In this section, we provide a detailed comparison of other existing mobile AI benchmarks as compared to the summary we provided in Section~\ref{Related work}. The key differences stem from the following:

\begin{itemize}
\item \textbf{Performance}: We support vendor SDKs. Other benchmarks can only support NNAPI and TFLite delegates (e.g., see results here: https://browser.geekbench.com/ml/v0/inference). While it is easier for developers to use NNAPI and TFLite delegates, they don't showcase the hardware's full capabilities. Table 3 shows performance varies by 10\% based on delegates. But depending on the chipset, performance between delegates and vendor SDKs can vary (sometimes) by 7x due to poor/buggy op implementations (eg. Fig 5~\cite{buch2021ai})! OEMs use vendor SDKs to speed up everyday apps like photos, camera etc. So OEMs want devices to perform well, and benchmarking vendor SDKs is essential. But this needs open-source + vendor support, which only the MLPerf Mobile benchmark supports.

\item \textbf{Accuracy}: In MLPerf Mobile, accuracy comes first. Performance is measured with respect to that minimum quality target. Table 1 shows the accuracy targets we set based on community and industry consensus. Our quality targets are all $>$93\% FP32 (Table 1). Other benchmarks show results for arbitrary accuracy targets that are less (if at all) meaningful, e.g. int8 models with 52\% of FP32 accuracy for object detection and 81\% of FP32 accuracy for image classification. Such models would not be deployed and will indeed mislead the industry. See e.g. https://browser.geekbench.com/ml/v0/inference/94248.

\item \textbf{Transparency}: MLPerf is the only transparent mobile ML benchmark. Geekbench ML, AI-Benchmark, etc. do not transparently provide and/or describe their summary-score/weighting rationale, model provenance, quantization method, and optimization techniques. This makes reproducibility nearly impossible, unless with MLPerf Mobile.

\item \textbf{Validity}: MLPerf has a results audit process (Section VI.B). MLPerf hires an independent auditor to review and replicate the results using the vendor-submitted code. Unlike MLPerf, none of the other benchmarks do this, nor do they make their code publicly available.

\end{itemize}

More specifically, here are the detailed comparsions against the benchmarks listed in Table~\ref{tab:compare}.

\textbf{Aitutu} employs vendor SDKs to implement image classification based on the Inception V3 neural network \cite{szegedy2015rethinking}, using 200 images as test data~\cite{antutu, Antutu2019}. The object-detection model is based on SSD-MobileNet \cite{howard2017mobilenets, liu_2016}, using a 600-frame video as test data. The benchmark score is a measure of speed and accuracy---faster results with higher accuracy yield a greater final score. However, \textit{it is a closed-source application which limits result transparency.}

\textbf{AI-Benchmark} performs an machine-learning-performance evaluation on mobile systems with AI acceleration that integrate HiSilicon, MediaTek, Qualcomm, Samsung, and UniSoc chipsets~\cite{ignatov2019ai}. It evaluates 21 deep-learning tasks, including inference speed, accuracy and stability. It runs pre-selected models with various bit widths (INT8, FP16, and FP32) on the CPU and open-source or vendor-proprietary TFLite delegates. Performance-report updates appear on a website \cite{AIbench} after each major release of TFLite/NNAPI and new SoCs with AI acceleration. However, \textit{vendors cannot provide optimized SDKs solutions as in MLPerf, which makes a big difference.}

\textbf{AImark} by Master Lu (Ludashi) \cite{AImark} is an Android and iOS application, and uses vendor SDKs to implement its benchmarks. It includes ResNet-34 \cite{he2015deep}, Inception V3 \cite{szegedy2015rethinking}, SSD-MobileNet \cite{howard2017mobilenets, liu_2016}, and DeepLab v3+ \cite{chen2018encoderdecoder}. The benchmark judges mobile-phone AI performance by evaluating recognition efficiency, and it provides a line-test score. However, \textit{it too is closed-sourced which limits transparency and adoptability.} 

\textbf{Android Machine Learning Test Suite (MLTS)} is part of the Android Open Source Project~\cite{API}. MLTS includes an app that allows us to test the latency and accuracy of quantized and floating-point TFLite models (e.g., MobileNet and SSD-MobileNet) against a subset of the Open Images Dataset \cite{kuznetsova_2020}. It contains \textit{tests to validate the behavior of the drivers in corner case conditions, not do performance benchmarking}.

\textbf{GeekBenchML} assesses mobile ML inference performance~\cite{geekbenchML}. It supports TFLite and NNAPI delegates, but it lacks supports for vendor SDKs backends which are important for OEM applications and unlocking the full SoC performance. Also, \textit{it lacks strict minimum accuracy  targets (e.g. Table~\ref{tab:task_models}), without which the performance results can be meaningless.}

\textbf{MLPerf Inference} is another industry-standard open benchmark~\cite{reddi2020mlperf}. MLPerf Mobile serves the smartphone industry (4~Billion devices), which presents its own unique challenges as discussed in Section~\ref{Benchmarking Challenges}. Our findings around ALP, NNAPI vs. vendor SDKs etc. are domain-specific and reveal insights that are unique to mobile device performance, which MLPerf Inference does not because \textit{it is a benchmark for server-scale ML deployments, not mobile.}

\textbf{Neural Scope} from National Chiao Tung University \cite{NNAPI1, Neuralscope} developed an NNAPI application supporting FP32 and INT8 precisions. The benchmarks comprise object classification, object detection, and object segmentation, including MobileNet v2 \cite{sandler2019mobilenetv2}, ResNet-50 \cite{he2015deep}, Inception V3, SSD-MobileNet \cite{howard2017mobilenets, liu_2016}, and ResNet-50 with atrous-convolution layers \cite{chen2017deeplab}. Users can run the app on their mobile devices and immediately receive a cost/performance comparison, but \textit{it lacks the open-source transparency that is direly needed.}

\textbf{TensorFlow Lite} provides a utility to measure the latency of any TFLite model \cite{tensorflowlite}. A wrapper API is also available to reference how these models perform when embedded in an Android application. Users can select the NNAPI delegate, and they can disable NNAPI in favor of a hardware-offload back end. \textit{It is focused on benchmarking individual TFLite operators' performance, not ML tasks with rules and metrics.}

% \textbf{Geekbench.} Primate Labs created Geekbench \cite{PrimateLab, geekbench}, a cross-platform CPU-compute benchmark that supports Android, iOS, Linux, macOS, and Windows. The Geekbench 5 CPU benchmark features new applications, including augmented reality and machine learning, but lacks heterogeneous-IP support. Users can share their results by uploading them to the Geekbench Browser.

\textbf{UL Procyon AI Inference Benchmark} from UL Benchmarks, VRMark \cite{benchmarks, procyon} is an Android NNAPI CPU- and GPU-focused AI benchmark.  It contains MobileNet v3 \cite{Howard2019}, Inception V4 \cite{szegedy2015rethinking}, SSDLite MobileNet v3 \cite{Howard2019, liu_2016}, DeepLab v3 \cite{chen2018encoderdecoder}, and other models. It also attempts to test custom CNN models but uses an AlexNet \cite{NIPS2012_4824} architecture to evaluate basic operations. The application provides benchmark scores, performance charts, hardware monitoring, model output, and device rankings. But it \textit{only compares NNAPI implementations on FP and INT-optimized models, not other frameworks}.
 
\textbf{Xiaomi’s Mobile AI Benchmark} provides an end-to-end open-source tool for evaluating model accuracy and latency \cite{xianmiai}. The tool includes a daily performance-benchmark run for various neural-network models (mainly on the Xiaomi Redmi K30 Pro smartphone). The tool has a configurable back end that allows users to employ multiple ML-hardware-delegation frameworks (including MACE, SNPE, and TFLite). However, it is \textit{not designed for the broad ecosystem of different vendors' mobile devices}.

%%%%%%%%%%%%%%%%%%%%%%%%%%%%%%%%%%%%
\section{Future Work}
\label{Future work}

To enable the myriad use cases and reveal additional mobile processor insights, we are engaged in numerous activities.

\textbf{Expanding the benchmark suite} is an obvious area of improvement. We are working to expand the scope to include more tasks and models, along with different quality targets. Examples include additional vision tasks, such as super-resolution, as well on-device speech recognition~\cite{he2018streaming}. Our current network choices reflect common use cases, most mobile ML use cases involve computer vision. But our NLP Q\&A task \& Mobile BERT task are heavier. Speech RNN-T is in the works - we're working with Google and Facebook engineers to build a mobile model version.

Mobile has more heavy-duty models than our \textit{initial} selection. Super-resolution and high-resolution models are important use cases, but they are still evolving. However, there is no agreement on which mobile ML versions are broadly applicable for these types of use cases. Also, the metrics for evaluating these tasks are not clearly defined. The other issue is a lack of commercially distributable datasets for these tasks. Thus, to begin, we included reliable models that were stable and could be improved over time.

\textbf{End-to-end performance} is important as user-perceived latency includes often includes pre- and post-processing overheads, and it has been shown to be  non-negligible~\cite{aitaxmobile}. In the future, we may consider extending the scope of measurements.

\textbf{iOS support} recently became available. Example uses can be found online~\cite{MLPerfIn1:online}. Apple's iOS is a major AI-performance player that brings additional hardware and software diversity and we expect results in the near future.

\textbf{Measuring software frameworks} is essential. As we described in Section~\ref{Benchmarking Challenges}, software performance---and, more importantly, its capabilities---is crucial to unlocking a device's full potential. Enabling apples-to-apples comparison of software frameworks on a fixed hardware platform has merit. The back-end code path in Figure~\ref{fig:code path} (code path 1) is a way to integrate different machine-learning frameworks to determine which one achieves the best performance on a target device.

\textbf{Power measurement} is a major area of potential improvement. Since mobile devices are battery-constrained, evaluating mobile AI's power draw is important. While we currently do not account for power, most smartphone chipsets are capped at a 3W TDP, so it provides an artificial thermal/power ceiling.

\textbf{Rolling result submissions} are needed as new devices frequently arrive, often in between the calls for submissions. We plan to add ``rolling submissions'' to encourage vendors to submit their scores continuously, which would allow up-to-date and consistent reporting of the AI performance. Technical roadmaps like IRDS~\cite{irds} rely on this data to make informed recommendations to policymakers and funding agencies.

% To make additional progress, we need community involvement. We therefore encourage the broader mobile community to join the MLPerf effort and maintain the momentum behind an industry-standard open-source mobile benchmark.

\section{Artifact Evaluation}

The MLPerf Mobile App artifact we provide is a mobile app which provides latency and accuracy benchmarking capabilities for four mobile-targeted neural network models, evaluated on three computer vision tasks and one natural language processing task. The app can be run standalone to evaluate performance (``performance mode'') or optionally on full validation datasets to evaluate both accuracy and performance (``accuracy mode''). Additionally, smartphone vendors can provide optimized back-end implementations to take advantage of architecture-specific features to increase inference speed. Here, we provide instructions here for reproducing performance mode evaluations without a custom back-end. The app is open-source, and a pre-built APK is available to reviewers on request (Sec.~\ref{deliver}).

\subsection{Artifact Check-list (Meta-information)}

% {\em Obligatory. Use just a few informal keywords in all fields applicable to your artifacts
% and remove the rest. This information is needed to find appropriate reviewers and gradually 
% unify artifact meta information in Digital Libraries.}

{\small
\begin{itemize}
%   \item {\bf Algorithm: }
  \item {\bf Program: } MLPerf Mobile App
  \item {\bf Compilation: } Android Studio (version $>=$ 4.0), TensorFlow Lite
%   \item {\bf Transformations: }
  \item {\bf Binary: } mlperf\_app.apk
  \item {\bf Datasets: } Depends on ImageNet ILSVRC2012, COCO2017, ADE20K, and SQuAD 1.1, which are available for download online.
  \item {\bf Models: } Depends on MobileNetEdgeTPU, SSD-MobileNet v2, MobileDET\_SSD, Deeplabv3+ - MobileNetv2, MobileBERT, which are available for download online.
  \item {\bf Run-time environment: } Android 10.0+ (API 29)
  \item {\bf Hardware: } Android smartphones
  \item {\bf Run-time state: } On battery power, with sufficient ventilation and in a room temperature between 20-25C, with cooldown periods of 5 minutes between tests.
%   \item {\bf Execution: }
  \item {\bf Metrics: } Queries per second, and optionally, Top-1 accuracy, mAP, mIoU, F1 score
  \item {\bf Output: } Displayed numerical results
  \item {\bf Experiments: } On-device benchmark measurements should be within the allowed quality targets (Table 1) and within 5\% of vendor-reported metrics (Sec. VI-B).
  \item {\bf How much disk space required (approximately)?: } under 100MB
  \item {\bf How much time is needed to prepare workflow (approximately)?: } 2 hours
  \item {\bf How much time is needed to complete experiments (approximately)?: } Under 1 hour
  \item {\bf Publicly available?: } The source code for the Android APK is archived here: [\url{https://doi.org/10.5281/zenodo.6416132}]. The source code is also available on github at [\url{https://github.com/mlcommons/mobile_app_open/tree/android-v2}]. Build artifact is distributed only to reviewers.
  \item {\bf Code licenses (if publicly available)?: } \url{https://github.com/mlcommons/mobile_app_open/blob/master/LICENSE.md}
%   \item {\bf Data licenses (if publicly available)?: }
%   \item {\bf Workflow framework used?: }
%   \item {\bf Archived (provide DOI)?: } To be provided in the camera-ready materials
\end{itemize}
}

%%%%%%%%%%%%%%%%%%%%%%%%%%%%%%%%%%%%%%%%%%%%%%%%%%%%%%%%%%%%%%%%%%%%%
\subsection{Description}

\subsubsection{Build Instructions}
\label{deliver}

Instructions for building the MLPerf Mobile App are provided at \url{https://github.com/mlcommons/mobile_app_open/tree/android-v2}, and in the archived version of this branch at \url{https://doi.org/10.5281/zenodo.6416132}. For convenience, we also have provided prebuilt Android APKs for Artifact Evaluation reviewers to use. Reviewers can email William Chou (\href{mailto:wchou@qti.qualcomm.com}{wchou@qti.qualcomm.com}) and Wookie Hong (\href{mailto:jwookie.hong@mlcommons.org}{jwookie.hong@mlcommons.org}) to request access to the pre-built APK. 

\subsubsection{Hardware Dependencies}

Smartphones equipped with the MediaTek Dimensity 820, Samsung Exnyos 990, and Qualcomm Snapdragon 865+ were used in the evaluation. Some additional Android smartphone models are supported.

% \subsubsection{Software dependencies}

\subsubsection{Datasets}

For running benchmarks in ``performance mode'' no additional datasets are required to be downloaded, and all data needed is present within the APK. The following four datasets are used in the ``accuracy mode'' benchmark evaluation: ImageNet ILSVRC2012, COCO2017, ADE20K, and SQuAD 1.1. Instructions on obtaining and formatting the datasets are available at \url{https://github.com/mlcommons/mobile_app_open/blob/android-v2/android/cpp/datasets/README.md} which is included as part of the source APK repository, and includes scripts for reformatting the datasets to the format expected by the MLPerf Mobile App.

%%%%%%%%%%%%%%%%%%%%%%%%%%%%%%%%%%%%%%%%%%%%%%%%%%%%%%%%%%%%%%%%%%%%%
\subsection{Installation}

For reviewers, we provide a prebuilt APK (Sec.~\ref{deliver}) which can be installed by transferring the APK to an Android phone or SDCard and selecting the APK through the file browser. To evaluate in ``performance mode'' only the APK itself is needed. For non-reviewers, instructions for building and installing the MLPerf Mobile App APK are provided at \url{https://github.com/mlcommons/mobile_app_open/tree/android-v2}. In particular there are multiple pathways documented for building the app, including via Bazel, Docker, and Android Studio. 
% %%%%%%%%%%%%%%%%%%%%%%%%%%%%%%%%%%%%%%%%%%%%%%%%%%%%%%%%%%%%%%%%%%%%%
% \subsection{Experiment workflow}

%%%%%%%%%%%%%%%%%%%%%%%%%%%%%%%%%%%%%%%%%%%%%%%%%%%%%%%%%%%%%%%%%%%%%
\subsection{Evaluation and Expected Results}

Open the MLPerf Mobile App and click the gear/settings icon and ensure ``Submission mode'' is toggled off (this ensures the app in ``performance mode'' where accuracy is not evaluated on the full validation datasets). ``Cooldown'' can be enabled to add to include a 5 minute pause betweeen benchmarks to avoid thermal throttling. Click back to exit the settings page, and scroll to the bottom to select the ``Test'' or ``Test Again'' button to begin the evaluation. 

Once the suite of benchmarks is completed, results (as queries per second) will be displayed for Image Classification, Object Detection, Image Segmentation, Language Processing, and Image Classification (offline). 

%%%%%%%%%%%%%%%%%%%%%%%%%%%%%%%%%%%%%%%%%%%%%%%%%%%%%%%%%%%%%%%%%%%%%
% \subsection{Experiment customization}

% %%%%%%%%%%%%%%%%%%%%%%%%%%%%%%%%%%%%%%%%%%%%%%%%%%%%%%%%%%%%%%%%%%%%%
% \subsection{Notes}

%%%%%%%%%%%%%%%%%%%%%%%%%%%%%%%%%%%%%%%%%%%%%%%%%%%%%%%%%%%%%%%%%%%%%
\subsection{Methodology}

Submission, reviewing and badging methodology:

\begin{itemize}
  \item \url{http://cTuning.org/ae/submission-20190109.html}
  \item \url{http://cTuning.org/ae/reviewing-20190109.html}
  \item \url{https://www.acm.org/publications/policies/artifact-review-badging}
\end{itemize}

%%%%%%%%%%%%%%%%%%%%%%%%%%%%%%%%%%%%%%%%%%%%%%%%%%%%%%%%%%%%%%%%%%%%%%%%%%%%%%%
%%%%%%%%%%%%%%%%%%%%%%%%%%%%%%%%%%%%%%%%%%%%%%%%%%%%%%%%%%%%%%%%%%%%%%%%%%%%%%%

\end{document}